\documentclass[letterpaper,10pt,conference]{ieeeconf}
\IEEEoverridecommandlockouts
\usepackage[T1]{fontenc}
\usepackage[utf8]{inputenc}
\usepackage[english]{babel}
\usepackage{cite}
\usepackage{amsmath,amssymb,amsfonts}
\usepackage{graphicx}
\usepackage{textcomp}
\usepackage{xcolor}
\usepackage{subcaption}
\usepackage{booktabs}
\usepackage{siunitx}
\usepackage{url}
\usepackage[breaklinks=true,colorlinks=false,hidelinks]{hyperref}

\newcommand{\smalltt}[1]{{\small\texttt{#1}}}
\newcommand{\footnotett}[1]{{\footnotesize\texttt{#1}}}

\def\BibTeX{{\rm B\kern-.05em{\sc i\kern-.025em b}\kern-.08em
    T\kern-.1667em\lower.7ex\hbox{E}\kern-.125emX}}

\title{\LARGE \bf
Learning Reusable Hybrid Motion Priors for Humanoid Locomotion from Motion Imitation
}

\author{Valerio Belli$^{1,2}$, Valerio Modugno$^2$, Enrico Mingo Hoffman$^3$, and Fabio Amadio$^3$%
\thanks{$^1$ The author is with Sapienza University of Rome, Rome, Italy.}%
\thanks{$^2$ The authors are with University College London, London, UK.}%
\thanks{$^3$ The authors are with Inria, Universit\'{e} de Lorraine, CNRS, Nancy,
France.}%
\thanks{Emails: \footnotett{belli.1849294@studenti.uniroma1.it}, \footnotett{v.modugno@ucl.ac.uk}, \footnotett{enrico.mingo-hoffman@inria.fr}, \footnotett{fabio.amadio@inria.fr}}%
\thanks{This work was supported by the ANR project MeRLin (ANR-24-CE33-0753-01) and by the INRIA-UCL Associated Team LEG-AI.}%
}

\begin{document}

\maketitle
\thispagestyle{empty}
\pagestyle{empty}

\begin{abstract}
Reinforcement learning can produce robust humanoid controllers, but each new
task is typically trained as a separate policy with its own reward design and
training process. Motion imitation provides an alternative source of
motor competence by training policies to track retargeted human motions, yet
the resulting controllers remain reference trackers and are not directly
usable as task policies. We propose a three-stage pipeline that turns
motion-imitation skills into a reusable hybrid motion prior (HMP) for humanoid
locomotion. First, an expert policy is trained to imitate retargeted human
motion-capture clips. Second, the expert is distilled into a frozen
architecture composed of a proprioceptive encoder, a residual
vector-quantized (RVQ) codebook, and an action decoder. Third, task-level policies are trained to solve
locomotion tasks by selecting discrete codebook entries while the HMP remains
frozen. We evaluate the method on velocity tracking, point-goal navigation,
and fall-recovery velocity tracking in simulation, and deploy the
velocity-tracking policy on a real Unitree G1 robot.
The distillation process preserves the tracking behavior of the expert, while
the resulting HMP can be reused without retraining as the action interface for
different downstream locomotion policies. The learned HMP reveals an
interpretable codebook structure in which the number of active RVQ stages
modulates the available gait patterns. We further show that training the
codebook with the rotation trick improves latent organization and reduces
downstream falls compared with a standard straight-through estimator.
Project website: \href{https://hucebot.github.io/hmp-project/}{\smalltt{https://hucebot.github.io/hmp-project/}}.
\end{abstract}

\section{Introduction}

Reinforcement learning (RL) has become a standard tool for legged and humanoid
control. Large-scale simulation makes it possible to train robust controllers
that can be deployed on physical robots~\cite{rudin2022walk,radosavovic2024real}.
Recent humanoid systems have further extended humanoid robot capabilities to
expressive whole-body control~\cite{cheng2024expressive,fu2024humanplus}.
Despite this progress, standard RL controllers are usually trained as
task-specific solutions. A new objective typically requires a new reward design
and a new training run, during which the policy has to reacquire basic motor
competence such as balance, stepping, contact timing, and whole-body
coordination. 

Motion imitation offers a systematic way to acquire such competence from data.
Given retargeted human motions, RL can train a humanoid to track reference
trajectories and thereby learn diverse natural skills without designing a
separate task reward for each behavior~\cite{peng2018deepmimic,liao2025beyondmimic}.
The resulting policies, however, are reference trackers: they execute a provided
motion reference, but they do not directly solve downstream tasks such as
following a command, reaching a goal, or reacting to task-level observations.
We address this gap by making the motor skills acquired by imitation available
through a reusable hybrid motion prior and training task-level policies to act
through that prior.

We study this problem in the context of humanoid locomotion on the Unitree G1,
using a three-stage training pipeline. First, a motion-imitation
expert is trained to track a diverse set of retargeted locomotion clips,
including different gaits, styles, velocities, and recovery motions. Second,
the expert is distilled into a
hybrid motion prior (HMP) composed of a proprioceptive encoder, a residual
vector-quantized (RVQ) codebook, and an action decoder. The
discrete component builds on vector-quantized representations, using residual
quantization to convert imitation latents into discrete choices available to
downstream policies. The codebook is trained with the rotation
trick~\cite{fifty2025rotation} to
improve gradient propagation through the discrete bottleneck and obtain a
better-structured discrete representation. Third, task-level policies are
trained by selecting discrete codebook entries while the
HMP remains frozen.

We evaluate the proposed pipeline on three locomotion tasks: velocity-command
tracking, point-goal navigation, and fall-recovery velocity tracking. Beyond
task performance, we analyze how the learned codebook expresses different
motion patterns, study the effect of
the number of active RVQ stages and the rotation trick, and deploy the
velocity-tracking policy on the real Unitree G1 humanoid.

To summarize, the main contributions of this work are:
\begin{itemize}
    \item a continuous-discrete HMP architecture
    that combines a proprioceptive latent with an RVQ codebook and
    enables task policies to reuse motion-imitation skills through code
    selection;
    \item a three-stage training procedure that first learns a motion-imitation
    expert, then distills it into the HMP using residual quantization and the
    rotation trick, and finally trains task policies while keeping the HMP
    frozen;
    \item an experimental validation on Unitree G1 locomotion showing that the
    same frozen HMP can be reused for velocity tracking, point-goal navigation,
    and fall-recovery velocity tracking, with analyses of the number of active
    RVQ stages, the rotation trick, and zero-shot sim-to-real deployment.
\end{itemize}

\begin{figure*}[t]
    \centering
    \includegraphics[width=0.96\textwidth]{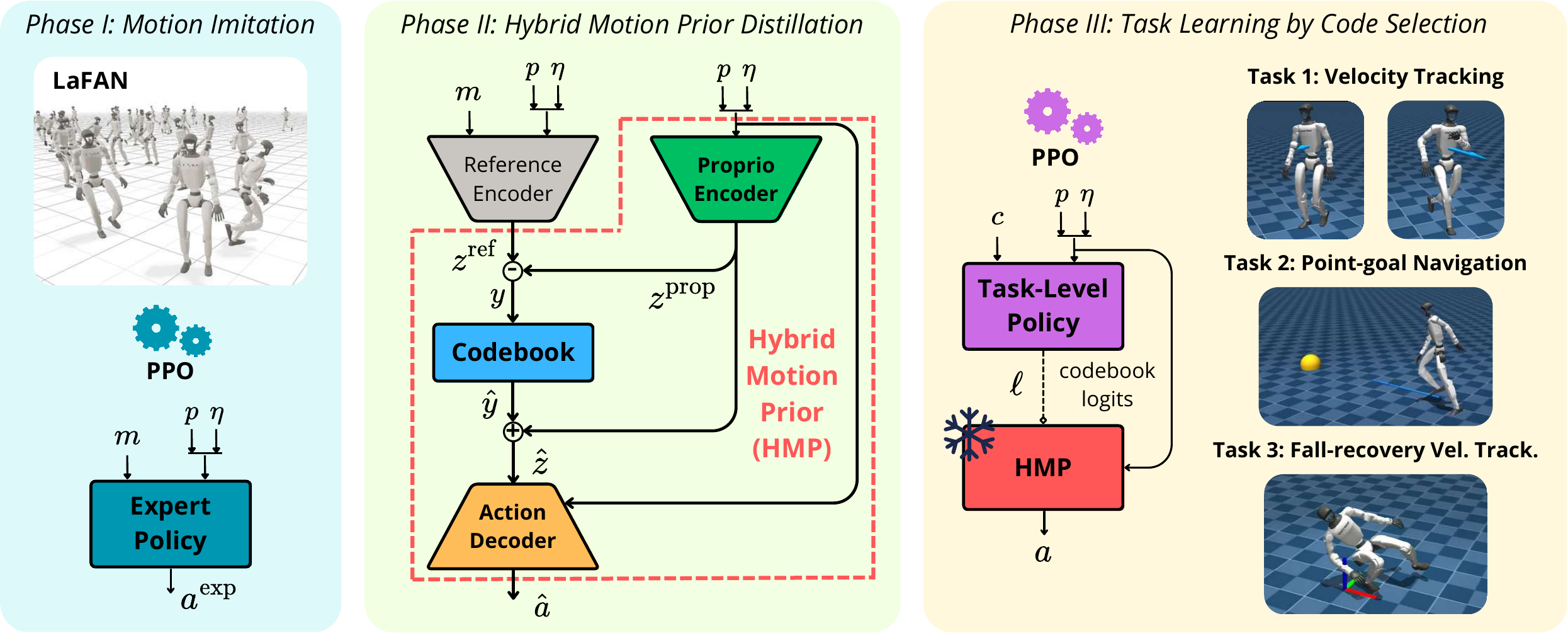}
    \caption{Overview of the three-stage pipeline. Phase I trains a
    motion-imitation expert to track retargeted mocap reference motions taken
    from the LaFAN dataset, using proprioception $p$, proprioceptive history
    embedding $\eta$, and motion command $m$. Phase II distills expert rollouts
    into the hybrid motion prior (HMP). During distillation, a training-only
    reference encoder receives $(p,\eta,m)$ and produces the reference latent
    $z^{\mathrm{ref}}$, while the deployable proprioceptive encoder receives
    $(p,\eta)$ and produces $z^{\mathrm{prop}}$. The codebook quantizes the
    residual $y=z^{\mathrm{ref}}-z^{\mathrm{prop}}$ into $\hat{y}$, and the
    action decoder reconstructs the expert action from
    $z^{\mathrm{prop}}+\hat{y}$. Phase III discards the reference encoder and
    freezes the HMP, composed of the proprioceptive encoder, codebook, and
    action decoder. A task-level policy then maps the task command $c$, $p$,
    and $\eta$ to codebook logits $\ell$, selecting codes that condition the
    frozen HMP. We evaluate this interface on three downstream locomotion tasks.}
    \label{fig:scheme}
\end{figure*}

\section{Related Works}
\subsection{Motion Imitation}

Motion imitation formulates whole-body control as reference tracking. DeepMimic
showed that example-guided RL can synthesize physics-based character skills
from motion clips~\cite{peng2018deepmimic}. Subsequent methods learned more
general humanoid tracking policies, improved robustness and scalability, and
used richer motion representations~\cite{luo2023phc,liao2025beyondmimic}.
The same tracking paradigm also supports expressive whole-body
control~\cite{cheng2024expressive,fu2024humanplus}.
These works establish reference tracking as a practical way to acquire humanoid motor skills from motion data. Our work starts from such a
tracker and studies how to reuse the acquired skills when no reference motion is
provided.

\subsection{Motion Priors for Reusable Control}

Prior work has explored reusable low-level control spaces in which downstream
policies query a learned controller or latent representation instead of acting
directly in the full joint-action space. Neural motor primitives provide an
early example~\cite{merel2019npmp}. Subsequent work developed reusable latent
control spaces for physics-based characters~\cite{peng2022ase,luo2024pulse} and
legged robots~\cite{11203023}.

Most of these priors are continuous. Discrete representations provide a
complementary interface by restricting task-level decisions to a finite
vocabulary of learned motor choices~\cite{11244136}. Vector quantization~\cite{vandenoord2017vqvae} and residual
vector quantization~\cite{lee2022rqvae} provide the
representation tools for this interface. Neural categorical
priors~\cite{zhu2023ncp} and hybrid continuous-discrete formulations based on
residual vector-quantized latents~\cite{bae2025hybrid} have shown strong
results in physics-based character control.

However, such methods have been explored mainly on simulated characters or
avatars, while real-robot applications have appeared primarily in fixed-base
robotic manipulation tasks~\cite{wu2025discretepolicy}. There are comparatively
few demonstrations of codebook-based action interfaces deployed on floating-base
humanoid robots.

\section{Method}

The proposed pipeline consists of three phases. Phase 1 trains a motion-imitation expert
from retargeted clips. Phase 2 distills this expert into a hybrid motion prior
(HMP). Phase 3 freezes the HMP and trains task-level policies that act by
selecting codebook indices.
The overall approach is illustrated in Fig.~\ref{fig:scheme}.

Let $p_t$ be the instantaneous proprioceptive observation available to the
policy at time $t$:
\begin{equation}
p_t =
\left[
{}^b\omega_t,\ {}^bg_t,\ q_t-q^{\mathrm{nom}},\ \dot{q}_t,\ a_{t-1}
\right],
\end{equation}
where ${}^b\omega_t$ is the base angular velocity, ${}^bg_t$ is the
projected gravity vector, $q_t$ and $\dot{q}_t$ are the joint positions and
velocities, $q^{\mathrm{nom}}$ is the nominal joint posture, and $a_{t-1}$ is the
previous action. Actions $a_t$ are interpreted as position targets for the
joint-level PD controllers. Superscripts $b$ and $w$ denote quantities
expressed in the base and world frames, respectively.

\subsection{Phase 1: Motion-Imitation Expert}

The first phase trains a motion-imitation expert using the training setup
of~\cite{amadio2026yahmp}. At each environment reset, we
randomly sample a motion clip and an initial time index within that clip. At
each control step, the sampled reference is encoded by a motion command
\begin{equation}
m_t =
\left[
\hat{q}_t,\dot{\hat{q}}_t,
{}^b\hat{v}_{xy,t},
{}^b\hat{\omega}_{z,t},
{}^w\hat{h}_t,
{}^w\hat{\phi}_t,
{}^w\hat{\theta}_t
\right],
\end{equation}
where hatted quantities are extracted from the reference:
$\hat{q}_t$ and $\dot{\hat{q}}_t$ are joint positions and velocities,
${}^b\hat{v}_{xy,t}$ is the local base planar velocity,
${}^b\hat{\omega}_{z,t}$ is the local yaw rate, ${}^w\hat{h}_t$ is the base height,
and ${}^w\hat{\phi}_t,{}^w\hat{\theta}_t$ are the base roll and pitch angles.

The reward combines base, body, joint, and
velocity tracking terms with regularization on contacts, action smoothness, and
joint limits. Episodes terminate on severe tracking failures.
We use domain randomization and external perturbations to improve robustness to
observation and modeling errors.

The expert is optimized with PPO~\cite{schulman2017proximal} within an
asymmetric actor-critic framework.
The actor receives noisy measurements of $p_t$, the proprioceptive history, and
$m_t$, while the critic additionally receives privileged information, including
noise-free measurements and physical quantities available only in simulation.

\subsubsection{Expert Architecture}
In our implementation, the expert actor uses a temporal convolutional history
encoder, $H_{\alpha}(\cdot)$, which maps the history window
$[p_{t-1},\ldots,p_{t-H}]$, with $H=10$, into a
proprioceptive history embedding
\begin{equation}
    \eta_t = H_{\alpha}([p_{t-1},\ldots,p_{t-H}]),
\end{equation}
extracting short-term motion features useful under changing
contact and dynamics conditions~\cite{li2025reinforcement}. Finally, an
encoder-decoder policy maps $p_t$, $\eta_t$, and $m_t$ to the
continuous expert action $a_t^{\mathrm{exp}}$ used for distillation in Phase 2.

\subsection{Phase 2: Hybrid Motion Prior Distillation}

The second phase distills the motion-imitation expert into a reusable HMP. The
frozen expert generates short on-policy rollouts in the same tracking
environment. We store the proprioceptive measurements, motion command $m_t$,
expert action $a_t^{\mathrm{exp}}$, and an episode-termination mask, and update
a student model with its own history encoder by supervised learning on the
resulting closed-loop data.

\subsubsection{Proprioceptive Encoder, Reference Encoder, and Action Decoder}
The distilled architecture separates the motion reference from proprioception
through two encoders that share a common latent space, and uses an action
decoder to map this space to actions.

The reference encoder $E_{\phi}(\cdot)$ has access to the current
motion-reference command $m_t$, which we treat as privileged information
available only during training. Together with the proprioceptive observation
$p_t$ and the history embedding $\eta_t$, it produces a latent intent
\begin{equation}
    z_t^{\mathrm{ref}} = E_{\phi}(p_t,\eta_t,m_t),
\end{equation}
whereas the proprioceptive encoder $P_{\theta}(\cdot)$ outputs the latent
\begin{equation}
    z_t^{\mathrm{prop}} = P_{\theta}(p_t,\eta_t).
\end{equation}
Of the two latent encoders, only $P_{\theta}$ is retained at deployment; the
reference encoder is used only during distillation.

\subsubsection{Residual Quantization}
Instead of quantizing the full latent, we quantize only the residual between the
reference latent and the proprioceptive latent:
\begin{equation}
    y_t = z_t^{\mathrm{ref}} - \mathrm{sg}(z_t^{\mathrm{prop}}),
    \label{eq:residual}
\end{equation}
where $\mathrm{sg}(\cdot)$ denotes the stop-gradient operator. The
proprioceptive encoder captures what can be inferred from the robot's
proprioception, while the codebook encodes only the residual motor intent
required to track the reference. The resulting lower-variance quantity makes
the discrete codes easier to learn. The discrete module is implemented as an
$M$-stage residual vector quantizer (RVQ). Let
$\mathcal{C}^{(m)}=\{e^{(m)}_1,\ldots,e^{(m)}_K\}$ denote the codebook of the
$m$-th quantization stage, with $K$ code entries. The RVQ selects one code
entry from each stage and sums the selected code vectors:
\begin{equation}
    \hat{y}_t = \sum_{m=1}^{M} e^{(m)}_{k_t^{(m)}} .
    \label{eq:rvqsum}
\end{equation}
During distillation, quantizer dropout evaluates the sum in
Eq.~\eqref{eq:rvqsum} over a randomly selected prefix of $M'\leq M$ active
stages. This encourages earlier stages to capture coarse motion structure,
while subsequent stages progressively refine the residual.
The action decoder $D_{\psi}(\cdot)$ receives the quantized residual added back
to the proprioceptive latent,
\begin{equation}
    \hat{z}_t = \mathrm{sg}(z_t^{\mathrm{prop}}) + \hat{y}_t,
\end{equation}
and reconstructs the expert action,
\begin{equation}
    \hat{a}_t = D_{\psi}(p_t,\eta_t,\hat{z}_t).
\end{equation}
The decoder is observation-injected: every hidden layer receives $[p_t,\eta_t]$
alongside the latent stream, which makes the low-level module a conditional
action generator rather than an unconditional latent-to-action map.

\subsubsection{Distillation Objective}
The student is trained by distillation with the composite loss
\begin{equation}
\mathcal{L} =
\lambda_a \mathcal{L}_{a}
+ \lambda_m \mathcal{L}_{m}
+ \lambda_r \mathcal{L}_{r}
+ \lambda_q \mathcal{L}_{q}.
\label{eq:distillation_loss}
\end{equation}
The action term is the squared error between the predicted and expert actions,
\begin{equation}
    \mathcal{L}_{a} = \|\hat{a}_t-a_t^{\mathrm{exp}}\|_2^2.
\end{equation}
The magnitude term penalizes the correction the residual adds on top of the
proprioceptive latent,
\begin{equation}
    \mathcal{L}_{m} = \|\hat{z}_t-z_t^{\mathrm{prop}}\|_2^2,
\end{equation}
and its weight $\lambda_m$ is annealed upward during training so that the
residual is not over-constrained before the student can reconstruct expert
actions. The temporal regularizer discourages abrupt changes in
$z_t^{\mathrm{prop}}$ and $\hat{y}_t$ between consecutive steps of the same
episode,
\begin{equation}
    \mathcal{L}_{r} =
    \mu_t\left(
    \|z_t^{\mathrm{prop}}-z_{t-1}^{\mathrm{prop}}\|_2^2
    +
    \|\hat{y}_t-\hat{y}_{t-1}\|_2^2
    \right),
\end{equation}
where the mask $\mu_t$ is $0$ across an episode boundary (immediately after a
termination) and $1$ otherwise, so that resets are not penalized as motion
discontinuities. Let $r_t^{(m)}$ denote the residual entering the $m$-th active
quantization stage, with $r_t^{(1)}=y_t$ and subsequent residuals obtained after
subtracting the codes selected by the preceding stages. The commitment loss is
\begin{equation}
    \mathcal{L}_{q}
    = \frac{\beta}{M'}
    \sum_{m=1}^{M'}
    \left\|r_t^{(m)}
    - \mathrm{sg}\!\left(e_{k_t^{(m)}}^{(m)}\right)\right\|_2^2.
\end{equation}
Codebook entries are updated separately through exponential moving averages of
their assigned residuals.

\subsubsection{Codebook Training with the Rotation Trick}
Because nearest-neighbor quantization is non-differentiable, gradients cannot
be backpropagated through the quantizer directly. Let $u$ denote the residual
entering one quantization stage and $e$ its selected code. The standard
straight-through estimator copies the output gradient directly to the input,
\begin{equation}
    \frac{\partial\mathcal{L}}{\partial u}
    \approx
    \frac{\partial\mathcal{L}}{\partial e}.
\end{equation}
The rotation trick~\cite{fifty2025rotation} keeps the same discrete forward
pass, but applies
$T_{u\rightarrow e}=\frac{\|e\|}{\|u\|}R_{u\rightarrow e}$ during the backward
pass:
\begin{equation}
    \frac{\partial\mathcal{L}}{\partial u}
    \approx
    T_{u\rightarrow e}^{\top}
    \frac{\partial\mathcal{L}}{\partial e},
\end{equation}
where $R_{u\rightarrow e}$ rotates $u$ onto $e$. We apply this transformation
independently at each active RVQ stage.

\subsubsection{Hybrid Motion Prior}
After distillation, the reference encoder is discarded. The history encoder is
retained to compute $\eta_t$ from the proprioceptive history. The HMP itself
consists of the proprioceptive encoder, codebook, and action decoder. Given
$p_t$, $\eta_t$, and codebook indices
$(k_t^{(1)},\ldots,k_t^{(M)})$, different index selections condition the
action decoder on different motion modes, producing the joint-position action
\begin{equation}
    a_t =
    D_{\psi}\left(p_t,\eta_t, P_{\theta}(p_t,\eta_t)
    +
    \sum_{m=1}^{M} e^{(m)}_{k_t^{(m)}}\right).
    \label{eq:controller_action}
\end{equation}
The HMP is the reusable action interface used by task-level policies in Phase 3.

\subsection{Phase 3: Downstream Task Learning by Code Selection}
The third phase trains a task-level policy while keeping the history encoder
and HMP frozen. The downstream task no longer provides a reference motion
$m_t$; instead it supplies a task-specific command $c_t$, which is passed to
the task-level policy together with $p_t$ and the history embedding $\eta_t$.

\subsubsection{Task-Conditioned Code Policy}
The task-level policy $F_{\xi}(\cdot)$ receives $p_t$, $\eta_t$, and $c_t$ and
outputs categorical logits for each quantization stage:
\begin{equation}
    \ell_t = F_{\xi}(p_t,\eta_t,c_t)
    \in \mathbb{R}^{M \times K}.
\end{equation}
For each stage, one index is sampled from the categorical distribution defined by
its logits,
\begin{equation}
    k_t^{(m)} \sim \mathrm{Cat}(\mathrm{softmax}(\ell_t^{(m)})),
    \quad m=1,\ldots,M .
\end{equation}
During evaluation and deployment, we instead select the highest-logit index at
each stage.
The resulting indices are used to retrieve code vectors from the frozen
codebook; the vectors are summed as in Eq.~\eqref{eq:rvqsum} and passed to the frozen
HMP in Eq.~\eqref{eq:controller_action}, which returns the joint-position action
applied to the robot.

Within the actor, only $F_{\xi}$ is trained in Phase 3. PPO optimizes the categorical
code-selection policy from the downstream task reward, while the environment is
stepped with the continuous action produced by the frozen HMP. In this
way, new task policies learn to select among the motion codes acquired during
imitation without modifying the low-level motor vocabulary.

\section{Experimental Results}
This section evaluates the proposed pipeline as a complete locomotion system.
All policies are trained in MuJoCo simulation~\cite{todorov2012mujoco} through
the \texttt{mjlab} GPU-accelerated training framework~\cite{zakka2026mjlab}.
The robot model is the 29-DoF Unitree G1, and all policies run at $50$\,Hz. We
first describe the shared training setup, then report the results obtained by
the task-level policies in simulation and on the real robot. The supplementary
video (available at \href{https://youtu.be/Af9qPkuPM3w}{\smalltt{https://youtu.be/Af9qPkuPM3w}}) provides qualitative results for the experiments presented in this
section, including all three downstream tasks and the real-robot deployment.
The internal structure of the learned HMP is analyzed separately in
Sec.~\ref{sec:codebook_analysis}.

\subsection{Training Details}

\subsubsection{Motion-Imitation Expert}
The expert is trained with PPO for $30{,}000$ iterations on a subset of 27
LaFAN motions~\cite{harvey2020robust} retargeted to the Unitree G1 morphology,
comprising 12 walking, 4 running, 2 sprinting, 3 jumping, and 6
fall-and-get-up motion clips. At reset, the robot is initialized from a sampled
reference state with randomized pose, base velocity, and joint-position
perturbations. Motion clips and phases are sampled adaptively, increasing the
probability of references and time segments that previously caused tracking
failures. During training, the expert is also exposed to observation noise,
randomized foot friction and center-of-mass offsets, and external velocity
perturbations applied every $1$--$3$\,s; self-collision contacts are enabled.

\subsubsection{Hybrid Motion Prior Distillation}
In the distillation stage, the proprioceptive history embedding, encoder
latents, and RVQ code vectors are all 128-dimensional. We use an eight-stage
RVQ ($M=8$) with $K=1{,}024$ code entries per stage. Codebook entries are updated with
exponential moving averages, with a decay of $0.99$. Quantizer dropout is
enabled during distillation, with $M'$ sampled uniformly from $\{1,\ldots,M\}$
at each update. The distillation loss weights are $\lambda_a=10$,
$\lambda_m=1$, $\lambda_r=0.05$, and $\lambda_q=1$, with commitment weight
$\beta=1$. The HMP is distilled for $6{,}000$ iterations. Once trained, the
same frozen HMP is reused for all downstream tasks.

\subsubsection{Downstream Task Policies}
Each downstream task is specified by its command space, reward, and command
sampling strategy. We evaluate three tasks.

\textbf{Velocity tracking task.}
The task-level policy receives desired planar and yaw velocities as task
commands and is rewarded for tracking them while maintaining stable locomotion.
Commands are resampled periodically during training, with a curriculum that
gradually expands from easy forward walking to omnidirectional velocity
tracking. The curriculum terminates with a challenging command range of
$v_x\in[-1.0,3.0]\,\mathrm{m\,s^{-1}}$,
$v_y\in[-1.0,1.0]\,\mathrm{m\,s^{-1}}$, and
$\omega_z\in[-3.0,3.0]\,\mathrm{rad\,s^{-1}}$.

\textbf{Point-goal navigation.}
The task-level policy receives a target position expressed relative to the robot
base and is rewarded for reducing the distance between the robot position and
the target while maintaining balance. A new goal is sampled whenever the robot
reaches the current one.

\textbf{Fall-recovery velocity tracking.}
The task-level policy receives the same velocity commands as in the standard
velocity-tracking task, but falling is treated as a recoverable state rather
than an episode failure. Training first exposes the policy to fallen reference
poses, used to initialize $30\%$ of episodes, and subsequently introduces
strong randomized base-velocity perturbations during locomotion. This
curriculum trains the policy to stand up, recover from induced falls, and
resume velocity tracking.

All downstream task policies are trained with PPO using 4,096 parallel
environments. Observation noise and dynamics randomization are retained, while
external perturbations are adapted to each task. Velocity tracking, point-goal
navigation, and fall-recovery velocity tracking are trained for 4,500, 5,500,
and 9,000 iterations, respectively.

\begin{figure}[t]
\centering

\begin{subfigure}[!ht]{0.24\columnwidth}
    \centering
    \includegraphics[width=\linewidth]{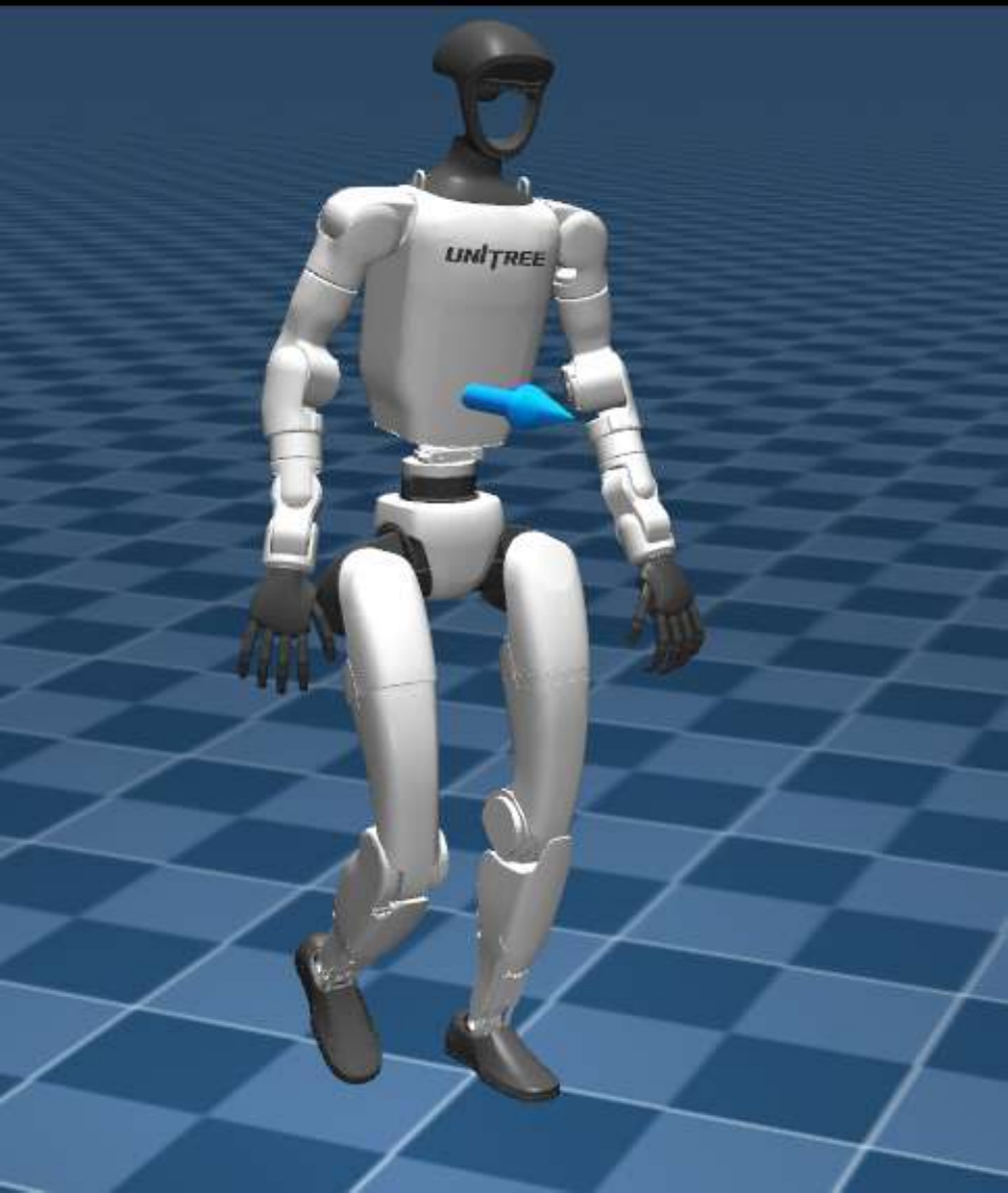}
    \caption{}
\end{subfigure}
\hfill
\begin{subfigure}[!ht]{0.24\columnwidth}
    \centering
    \includegraphics[width=\linewidth]{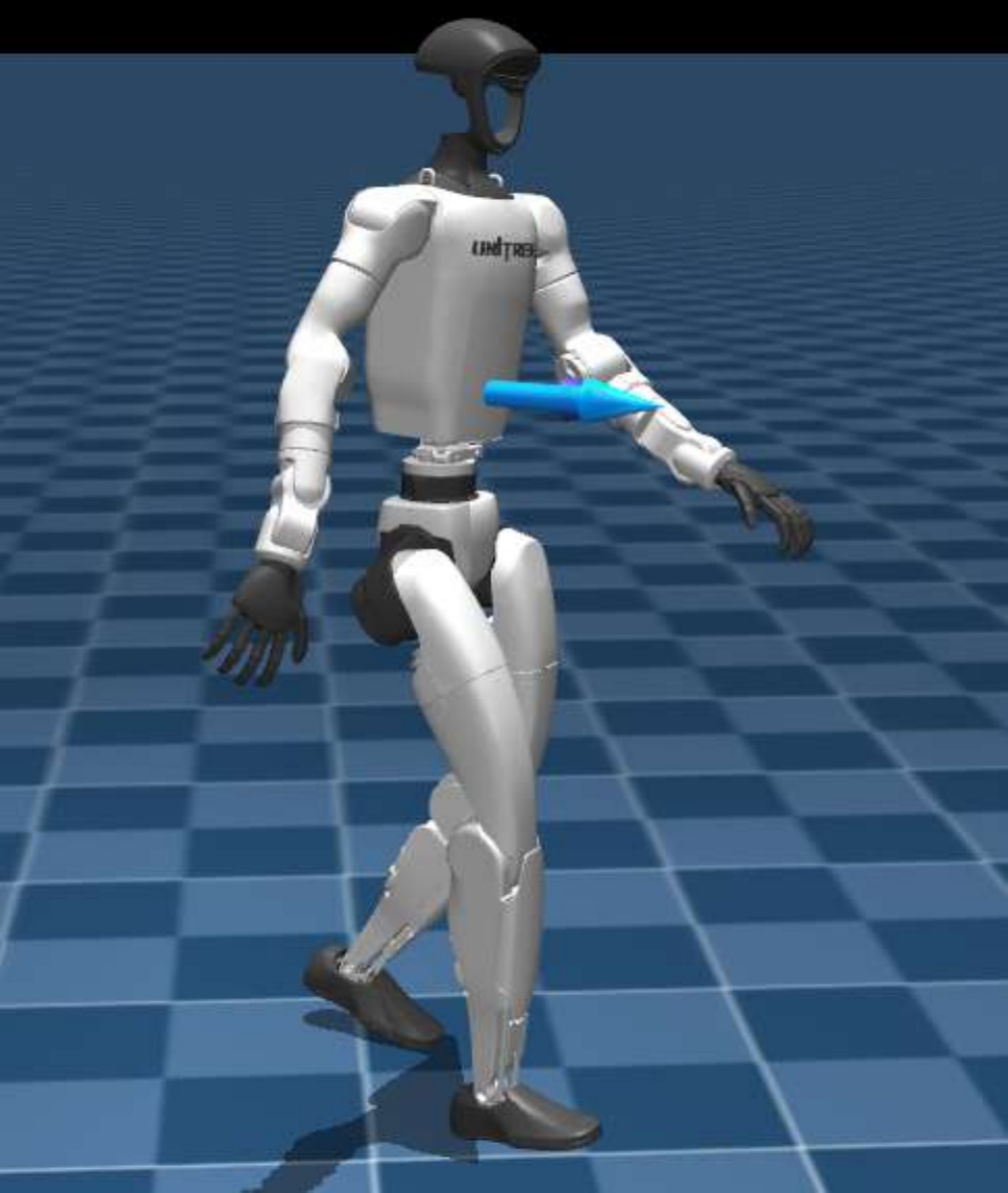}
    \caption{}
\end{subfigure}
\hfill
\begin{subfigure}[!ht]{0.24\columnwidth}
    \centering
    \includegraphics[width=\linewidth]{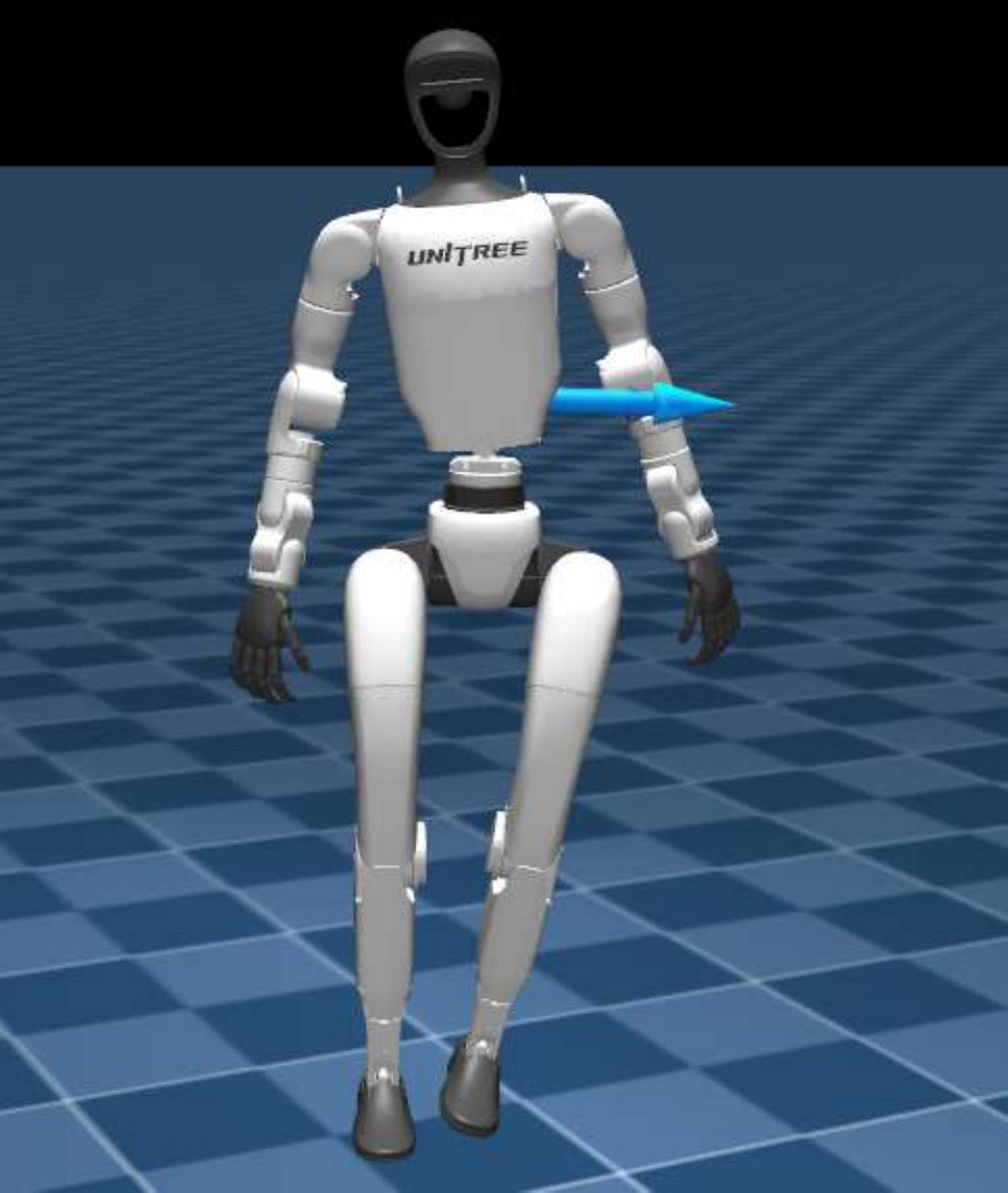}
    \caption{}
\end{subfigure}
\hfill
\begin{subfigure}[!ht]{0.24\columnwidth}
    \centering
    \includegraphics[width=\linewidth]{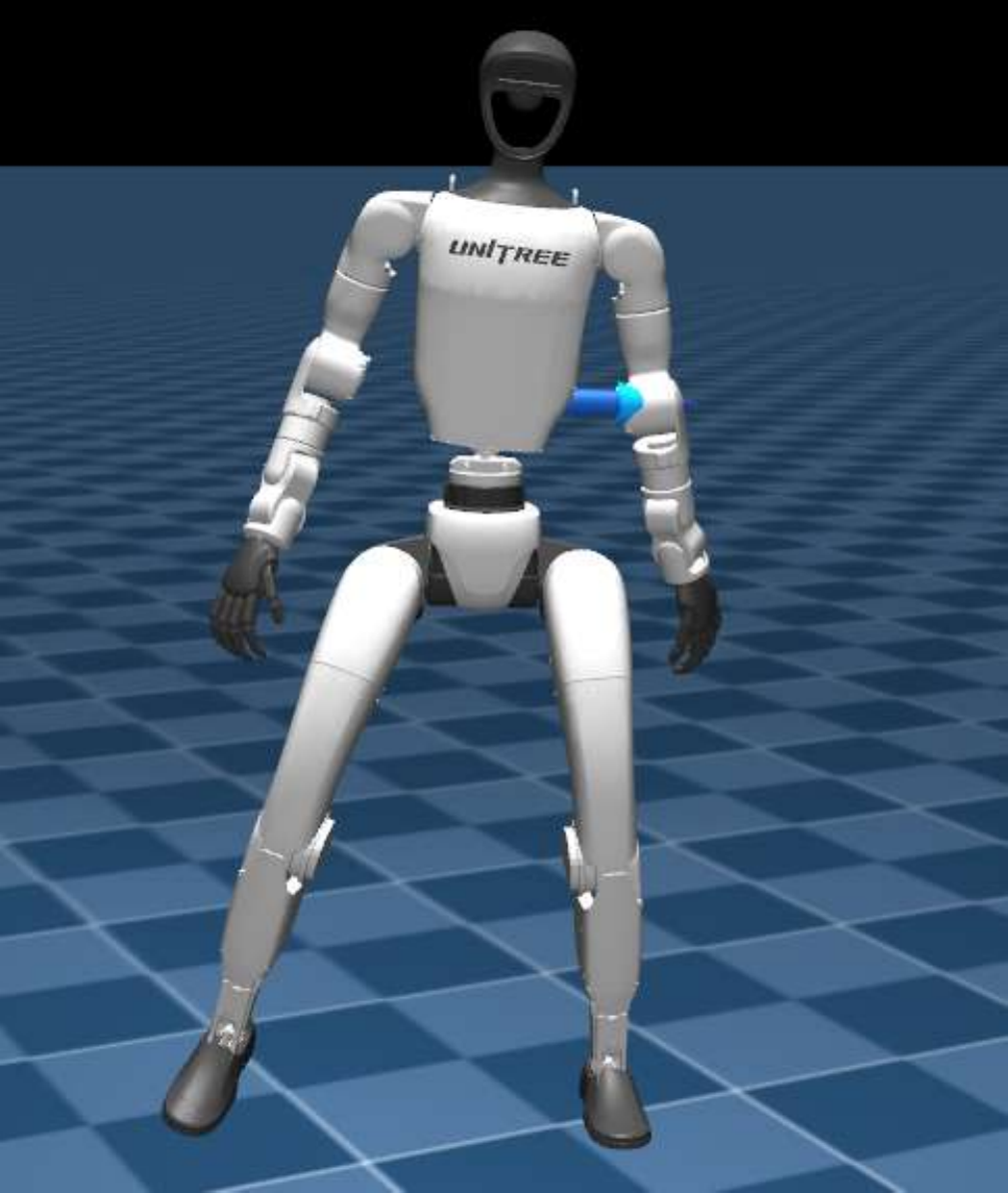}
    \caption{}
\end{subfigure}

\vspace{0.5em}

\begin{subfigure}[!ht]{0.24\columnwidth}
    \centering
    \includegraphics[width=\linewidth]{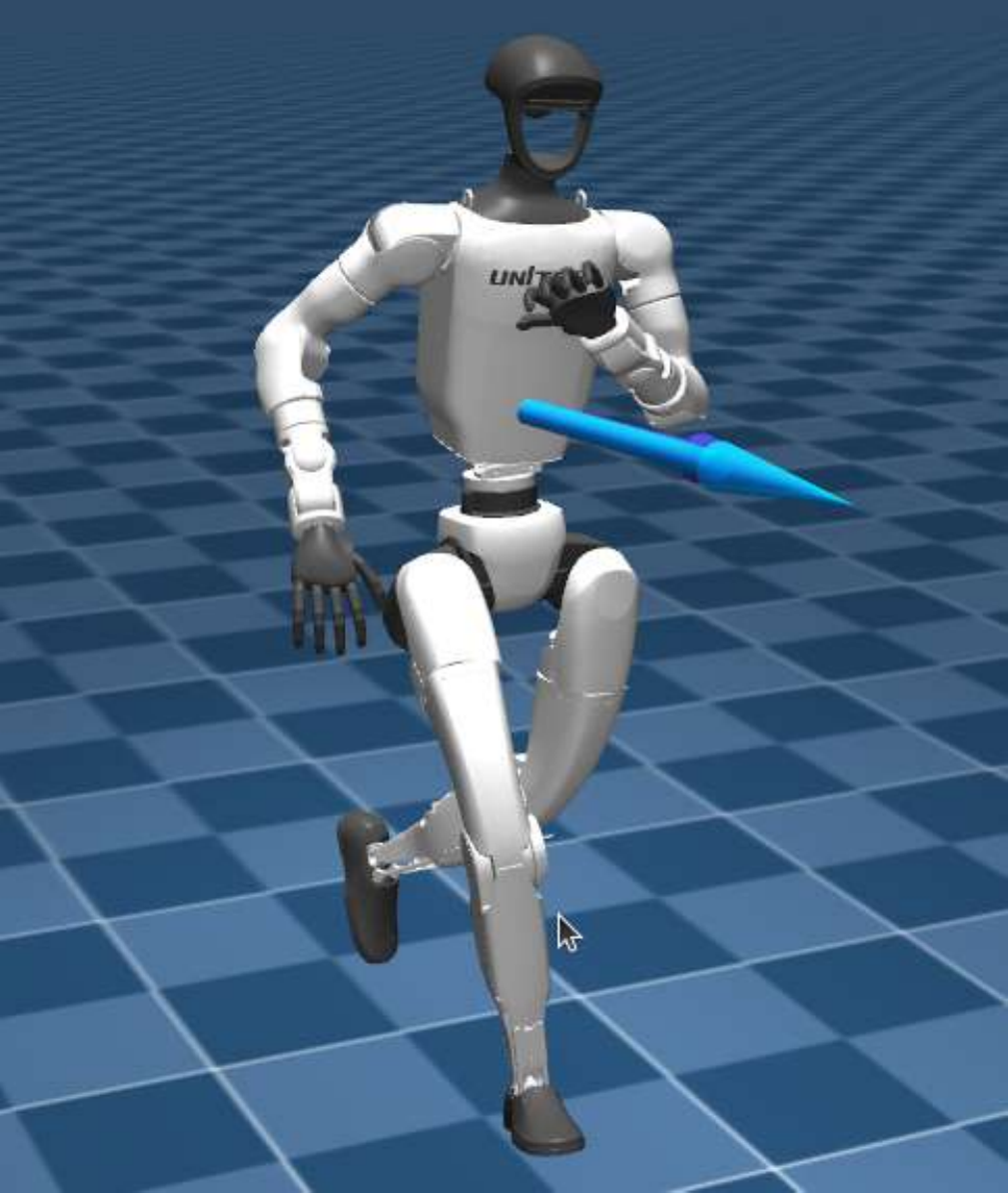}
    \caption{}
\end{subfigure}
\hfill
\begin{subfigure}[!ht]{0.24\columnwidth}
    \centering
    \includegraphics[width=\linewidth]{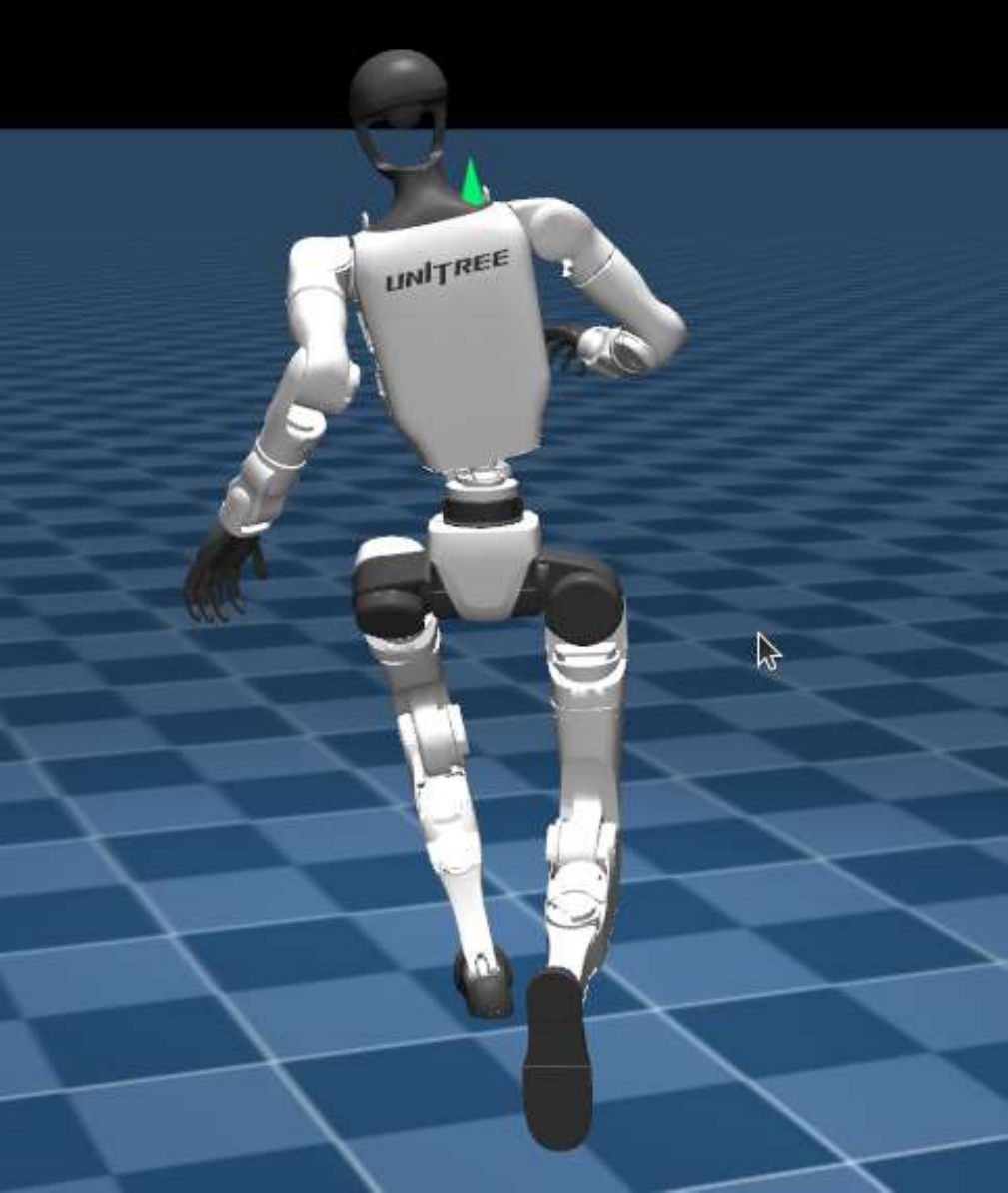}
    \caption{}
\end{subfigure}
\hfill
\begin{subfigure}[!ht]{0.24\columnwidth}
    \centering
    \includegraphics[width=\linewidth]{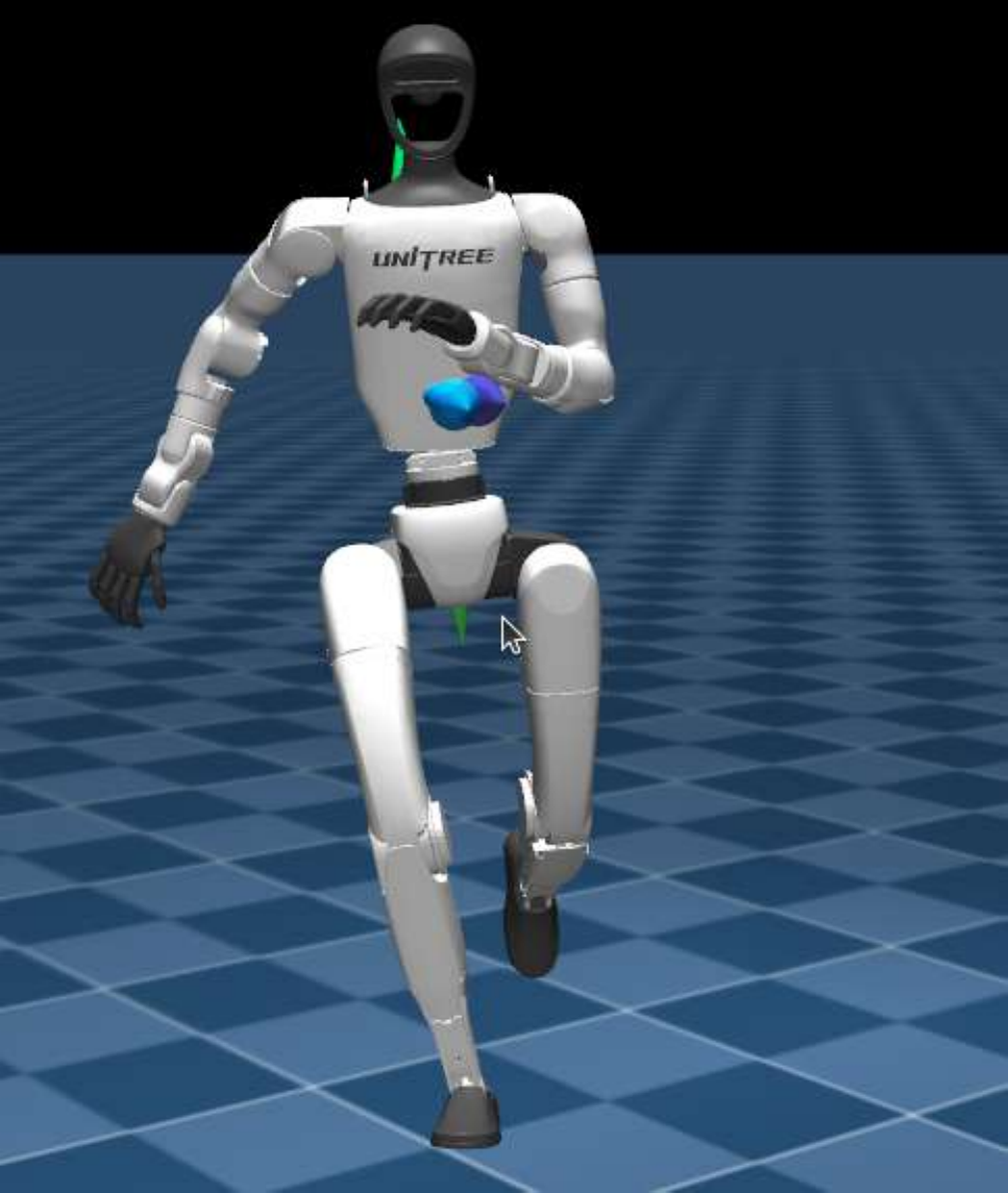}
    \caption{}
\end{subfigure}
\hfill
\begin{subfigure}[!ht]{0.239\columnwidth}
    \centering
    \includegraphics[width=\linewidth]{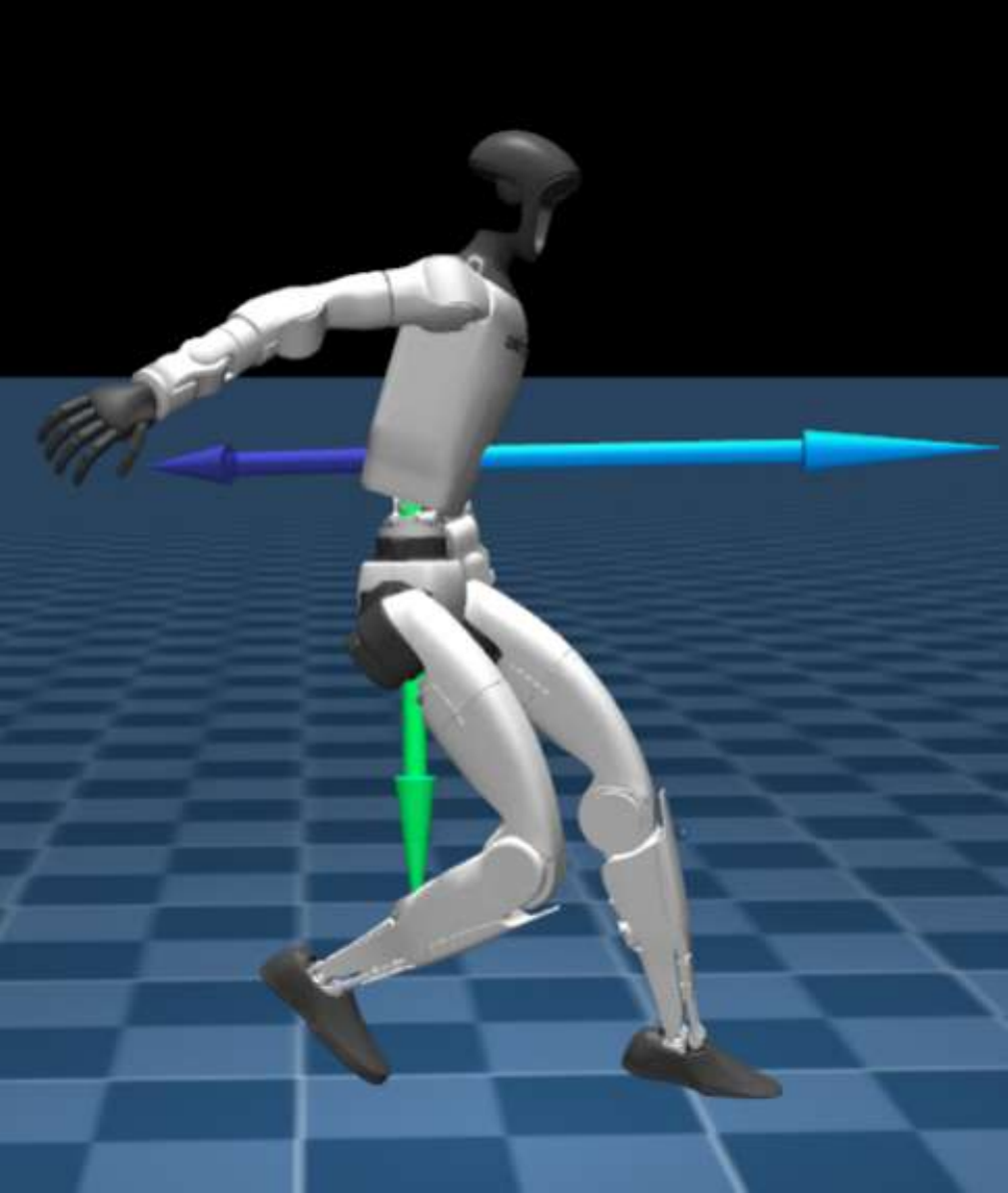}
    \caption{}
\end{subfigure}
\vspace{0.5em}

\caption{Representative velocity-tracking behaviors: walking (a--b),
sidestepping (c--d), running (e), turning (f--g), and a stopping maneuver (h)
under different velocity commands.}
\label{fig:frames}
\end{figure}

\subsection{Distillation Evaluation}
\label{sec:distillation_eval}

Before training downstream task policies, we verify that the distillation stage
preserves the behavior of the motion-imitation expert. We evaluate the expert
and the complete distilled model, comprising the reference encoder and the HMP,
on the same retargeted LaFAN motion set using $1{,}024$ simulation rollouts
of $30$\,s. During this evaluation, the reference encoder provides the
motion-conditioned latent used to compute the residual quantized by the RVQ. The
distilled model closely matches the
expert across the main tracking metrics: body-position error is $62.2$\,mm for
the distilled model and $62.4$\,mm for the expert, body-orientation error is
$0.210$\,rad versus $0.209$\,rad, and joint-position error is $0.721$\,rad
versus $0.712$\,rad. Action statistics are also nearly unchanged, with mean
action norm $7.25$ versus $7.20$ and mean action rate $0.93$ versus $0.92$.
These results confirm that distillation preserves the expert's closed-loop
tracking behavior before downstream code-selection policies are trained.

\subsection{Locomotion Tasks in Simulation}
\label{sec:locomotion_sim}

\textbf{Velocity tracking.}
The task-level policy reliably follows commanded planar and yaw velocities.
Over $1{,}024$ simulation rollouts of $30$\,s under the final command range, the
policy achieves a linear velocity RMSE of
$0.46\pm0.15\,\mathrm{m\,s^{-1}}$ and an angular velocity RMSE of
$1.03\pm0.46\,\mathrm{rad\,s^{-1}}$, with $21$ falls ($2.1\%$).
Qualitatively, the policy does not produce a single fixed gait: it switches
between standing, walking, fast walking, and running as the
commanded velocity changes, while also coordinating turns and lateral
commands.
Fig.~\ref{fig:frames} shows some illustrative examples.

\begin{figure}[t]
    \centering
    \includegraphics[width=\columnwidth]{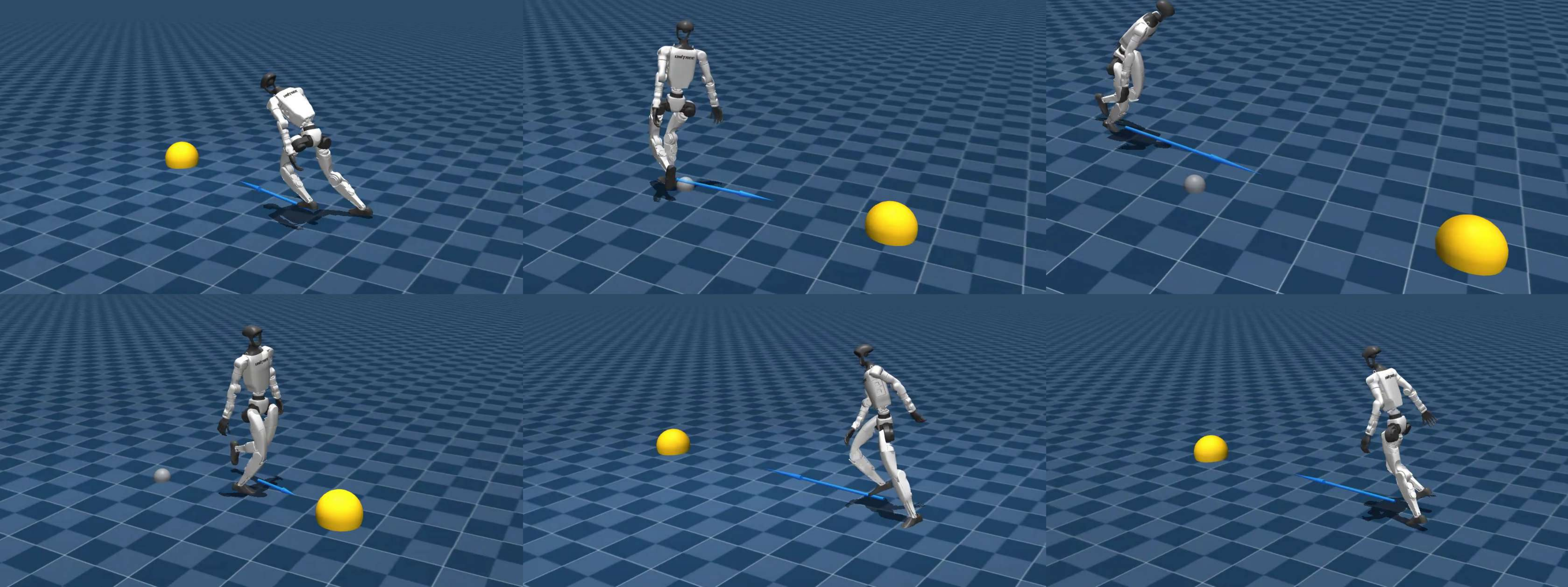}
    \caption{Example sequence from a point-goal navigation rollout. Yellow
    spheres indicate the active targets.}
    \label{fig:nav}
\end{figure}
\begin{figure}[t]
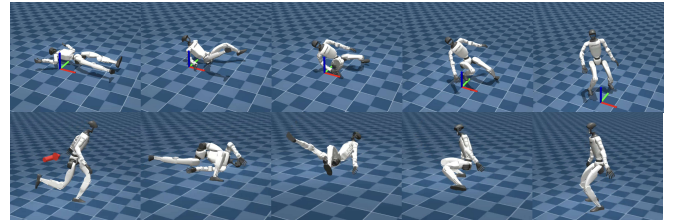

    \centering
    \begin{subfigure}{\columnwidth}
        \centering
        \includegraphics[width=\linewidth]{img/rec_1_strip.pdf}
    \end{subfigure}
    \begin{subfigure}{\columnwidth}
        \centering
        \includegraphics[width=\linewidth]{img/rec_2_strip.pdf}
    \end{subfigure}
    \caption{Stand-up from a lying configuration (top) and recovery after a
    strong external perturbation while running (bottom). The red arrow
    indicates the perturbation.}
    \label{fig:recovery}
\end{figure}

\textbf{Point-goal navigation.}
The policy moves through consecutive targets by combining goal-directed
walking, heading corrections, and gait adaptation. In a scripted evaluation
consisting of $1{,}024$ simulation rollouts with slalom, circular, and
high-curvature target sequences, the policy reaches a new goal roughly every
$1.4$\,s, moves at an average speed close to
$2\,\mathrm{m\,s^{-1}}$, and falls in $4.2\%$ of the trials.
Fig.~\ref{fig:nav} shows representative frames.

\textbf{Fall-recovery velocity tracking.}
We evaluate the policy under repeated external disturbances while tracking
velocity commands. Over $1{,}024$ simulation rollouts of $30$\,s, an
instantaneous base-velocity perturbation of $5\,\mathrm{m\,s^{-1}}$ along the
sagittal axis or $3\,\mathrm{m\,s^{-1}}$ along the lateral axis is applied
every $5$\,s.
The policy recovers from approximately $98\%$ of the induced falls, returns to
an upright posture in $1.5\pm0.6$\,s on average, and resumes velocity tracking.
Fig.~\ref{fig:recovery} shows representative stand-up and recovery sequences.

\subsection{Velocity Tracking on the Real Robot}

\begin{figure}[t]
    \centering
    \includegraphics[width=\columnwidth]{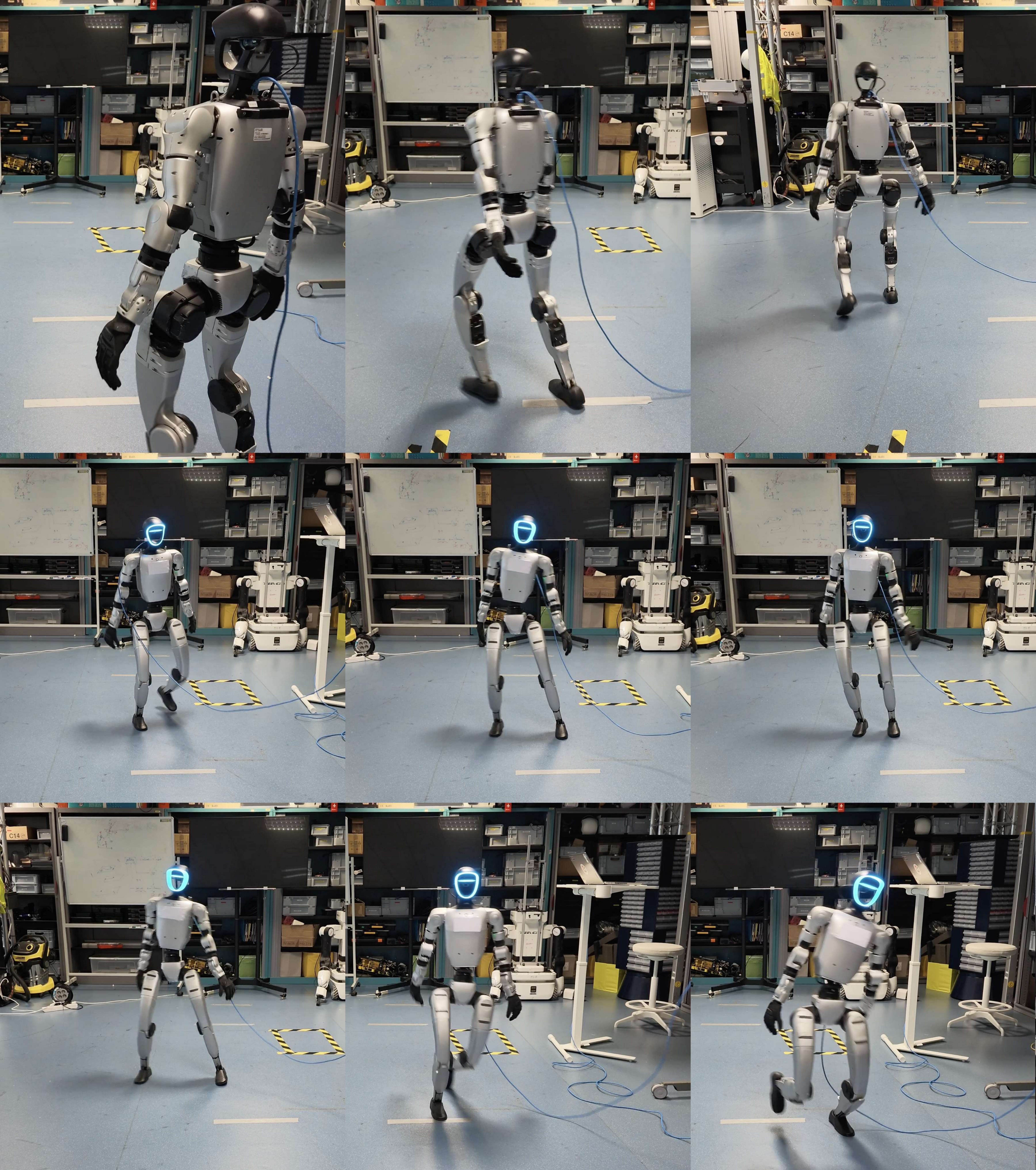}
    \caption{Zero-shot deployment on hardware. The velocity-tracking
    policy, transferred to a Unitree G1 without real-world fine-tuning,
    performs walking with turns and side-stepping over one continuous rollout.}
    \label{fig:real}
\end{figure}

We deploy the velocity-tracking policy on the Unitree G1 without additional
real-world fine-tuning. The robot tracks forward and lateral commands and
performs in-place and walking turns (Fig.~\ref{fig:real}).
The hardware behavior is qualitatively consistent with simulation. The robot
remains balanced and command-responsive, although visible oscillations make its
motion less smooth than the simulated rollouts.

\section{Analysis of the Hybrid Motion Prior}
\label{sec:codebook_analysis}
This section analyzes how the continuous proprioceptive encoder and discrete
codebook divide their roles, how active RVQ depth affects locomotion behaviors,
and how the rotation trick affects the organization of the learned codes. The
supplementary video also shows the behaviors discussed in this section.

\subsection{Role of the Proprioceptive Encoder and Codebook}
\label{sec:anatomy}

To separate the roles of the proprioceptive encoder $P_{\theta}$ and the
codebook, we evaluate the trained HMP while setting each contribution in
Eq.~\eqref{eq:controller_action} to zero in turn. First, the codebook
contribution is silenced ($\hat{y}_t=0$),
\begin{equation}
    a_t^{\mathrm{prop}}
    = D_{\psi}\!\left(p_t,\eta_t,\, z_t^{\mathrm{prop}}\right).
    \label{eq:prop_only}
\end{equation}
Second, the proprioceptive encoder is silenced ($z_t^{\mathrm{prop}}=0$),
\begin{equation}
    a_t^{\mathrm{cb}}
    = D_{\psi}\!\left(p_t,\eta_t,\, \hat{y}_t\right).
    \label{eq:cb_only}
\end{equation}

\subsubsection{Proprioceptive-encoder-only}
\begin{figure}[t]
    \centering
    \includegraphics[width=\columnwidth]{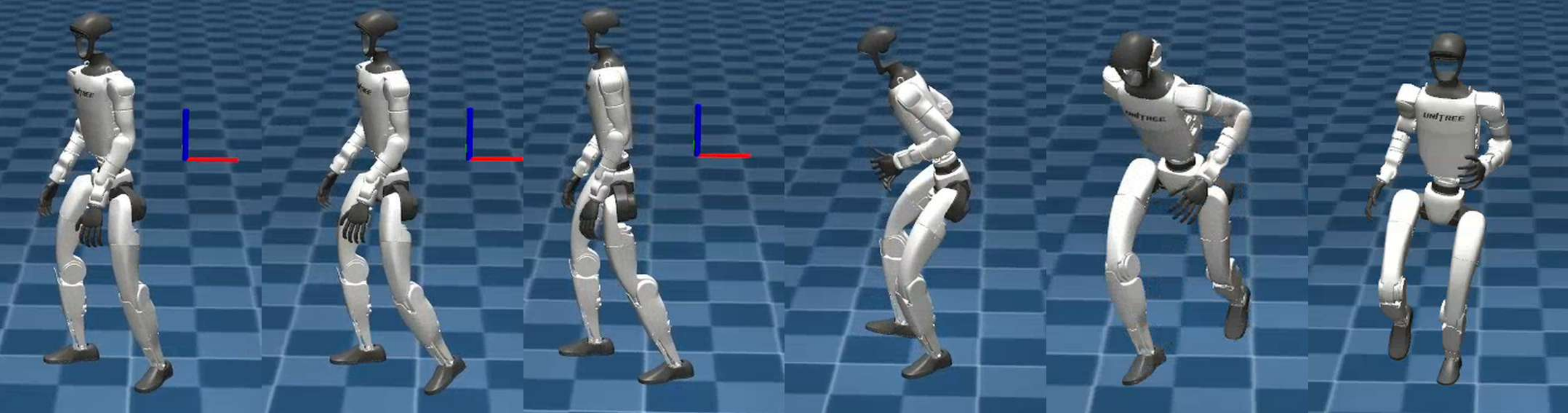}
    \caption{Proprioceptive-encoder-only rollout ($\hat{y}_t=0$). With the
    codebook silenced, the frozen HMP maintains balance and
    produces spontaneous locomotion patterns.}
    \label{fig:anatomy}
\end{figure}

With $\hat{y}_t=0$ the robot maintains static and dynamic balance, and spontaneously produces
locomotor patterns--standing, walking, and short running bursts--without
any goal-directed structure. This behavior suggests that $P_{\theta}$ captures
the balance and locomotion competence of the imitation dataset, but by
construction it has no access to a task command.
Fig.~\ref{fig:anatomy} shows a representative rollout.

\subsubsection{Codebook-only}
With $z_t^{\mathrm{prop}}=0$, the robot falls within a few control steps in
essentially every rollout. This is consistent with Eq.~\eqref{eq:residual}: the
codebook encodes modulations of the proprioceptive latent rather than complete
latent actions. Without $z_t^{\mathrm{prop}}$, the decoder receives an input
outside the distribution seen during distillation, and no stable motion is
produced.

These two tests support the intended split: the proprioceptive encoder acts
as a balance- and locomotion-aware backbone, while the codebook provides the
task-conditioned modulations selected by the task-level policy.

\subsection{RVQ Depth Analysis}

\label{sec:depth}
We next study how varying the number of active RVQ stages changes the
locomotion patterns available to the task-level policy. The RVQ contains eight
stages, and we define the \emph{active RVQ depth} $M_{\mathrm{act}}$ as the
number available to the task-level policy. We train one
policy for each active depth and report the representative settings
$M_{\mathrm{act}}\in\{1,2,3,5,8\}$; the omitted intermediate depths show
similar behavior. All other HMP components remain frozen.

Each policy is evaluated with a forward-speed progression. We run $1{,}024$ parallel
rollouts with a fixed world heading and commanded forward speeds
$v^\star\in\{0.0,0.5,1.0,1.5,2.0,2.5,3.0\}\,\mathrm{m\,s^{-1}}$.
Each speed is held for $5$\,s. The yaw-rate command is generated by a
proportional heading controller that keeps the robot aligned with the target
heading.
Table~\ref{tab:rvq_depth_locomotion} summarizes the results at each depth.

\begin{figure}[t]
    \centering
    \begin{subfigure}[t]{0.19\columnwidth}
        \centering
        \includegraphics[width=\linewidth,trim={26.46cm 6.00cm 26.46cm 6.00cm},clip]{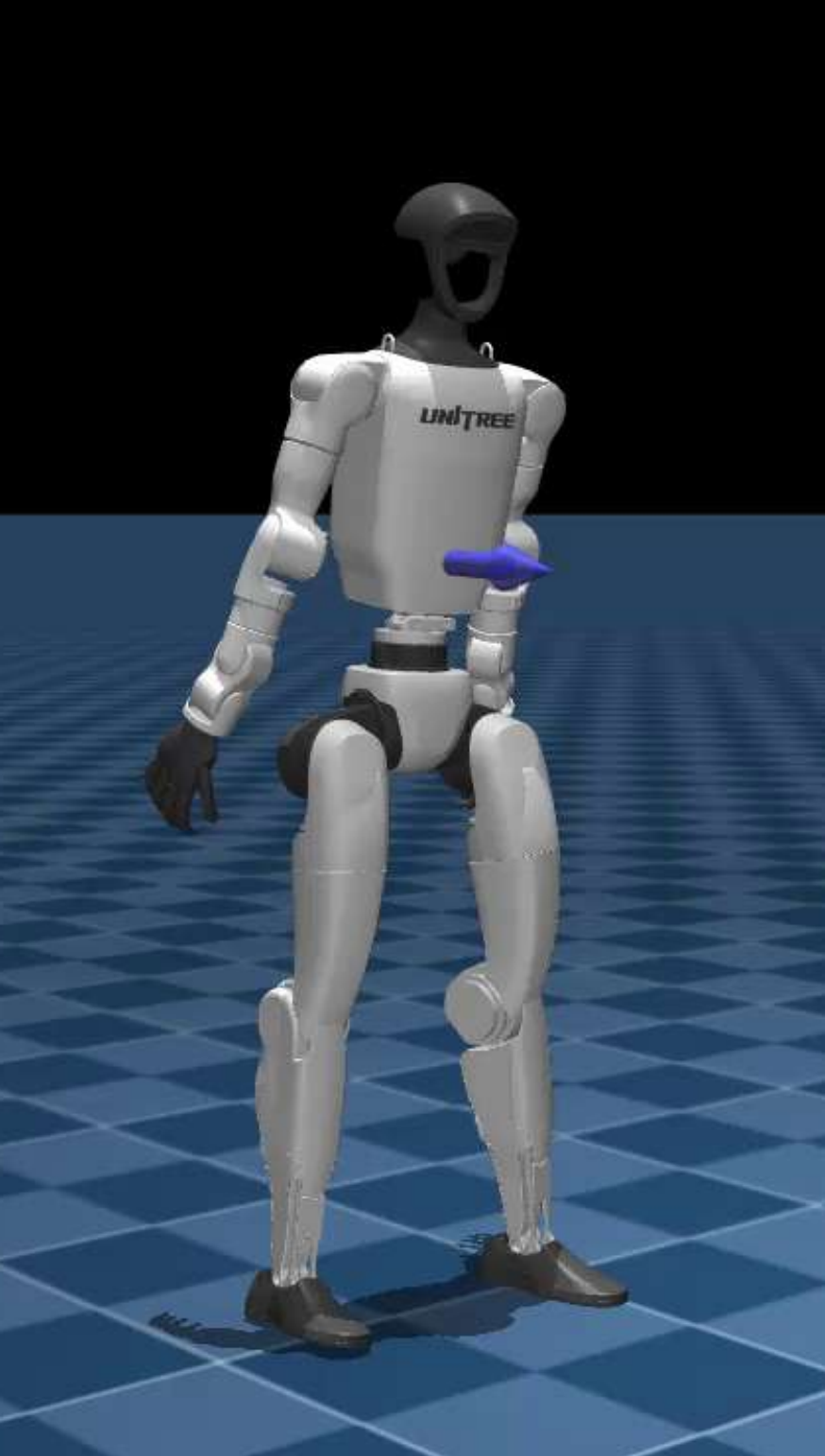}
        \caption{$1$}
    \end{subfigure}\hfill
    \begin{subfigure}[t]{0.19\columnwidth}
        \centering
        \includegraphics[width=\linewidth,trim={26.46cm 6.00cm 26.46cm 6.00cm},clip]{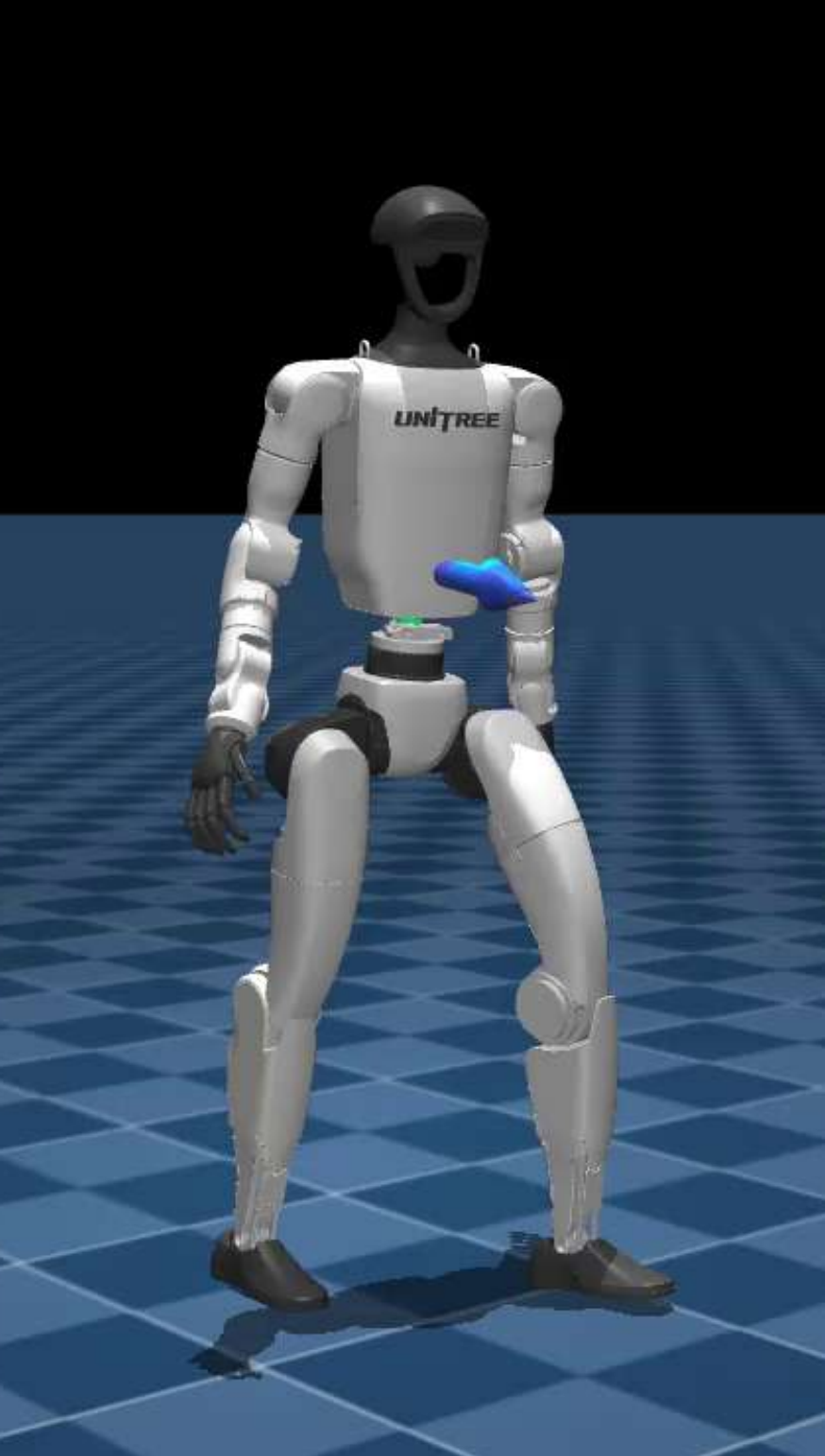}
        \caption{$2$}
    \end{subfigure}\hfill
    \begin{subfigure}[t]{0.19\columnwidth}
        \centering
        \includegraphics[width=\linewidth,trim={26.46cm 6.00cm 26.46cm 6.00cm},clip]{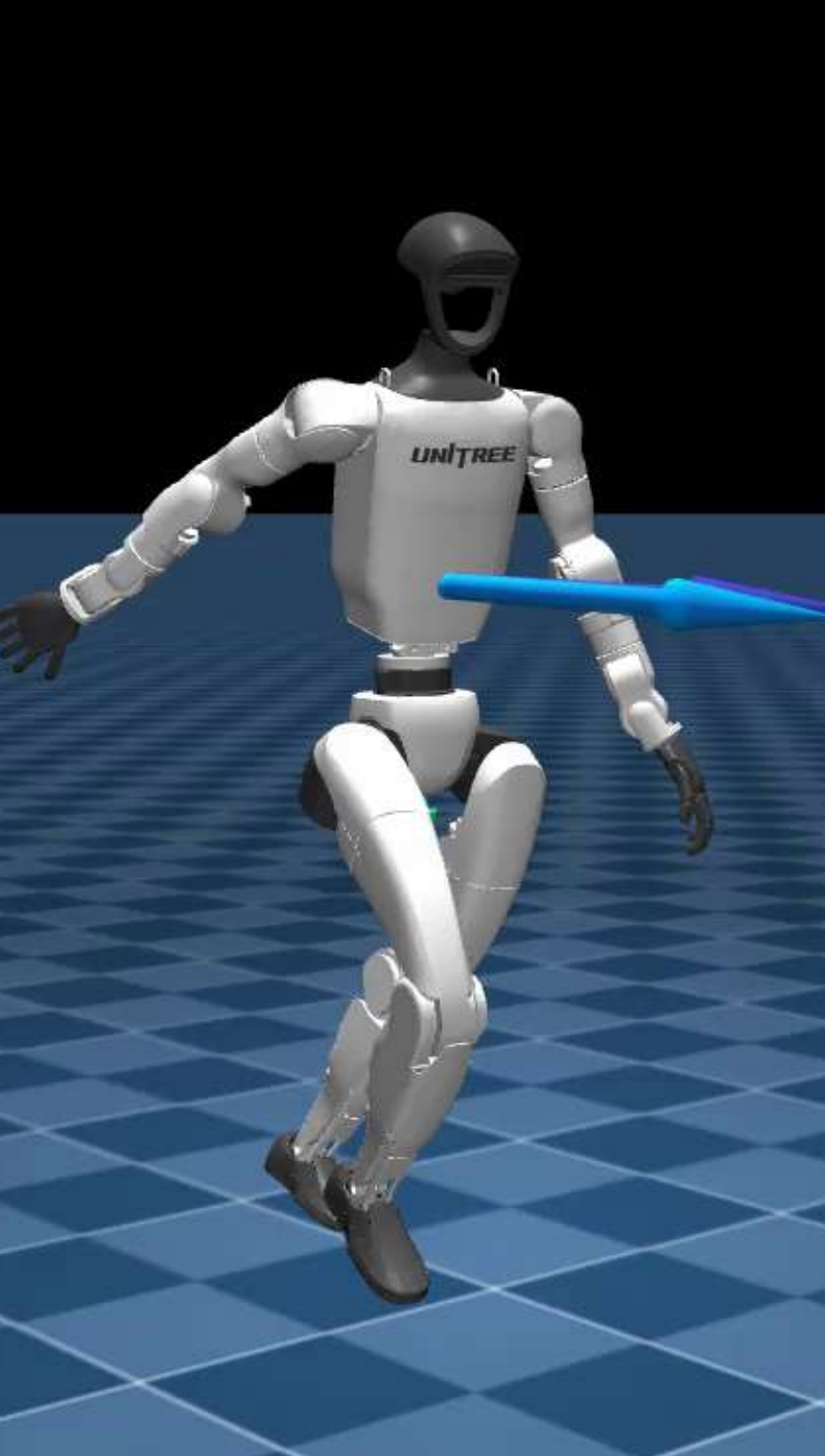}
        \caption{$3$}
    \end{subfigure}\hfill
    \begin{subfigure}[t]{0.19\columnwidth}
        \centering
        \includegraphics[width=\linewidth,trim={26.46cm 6.00cm 26.46cm 6.00cm},clip]{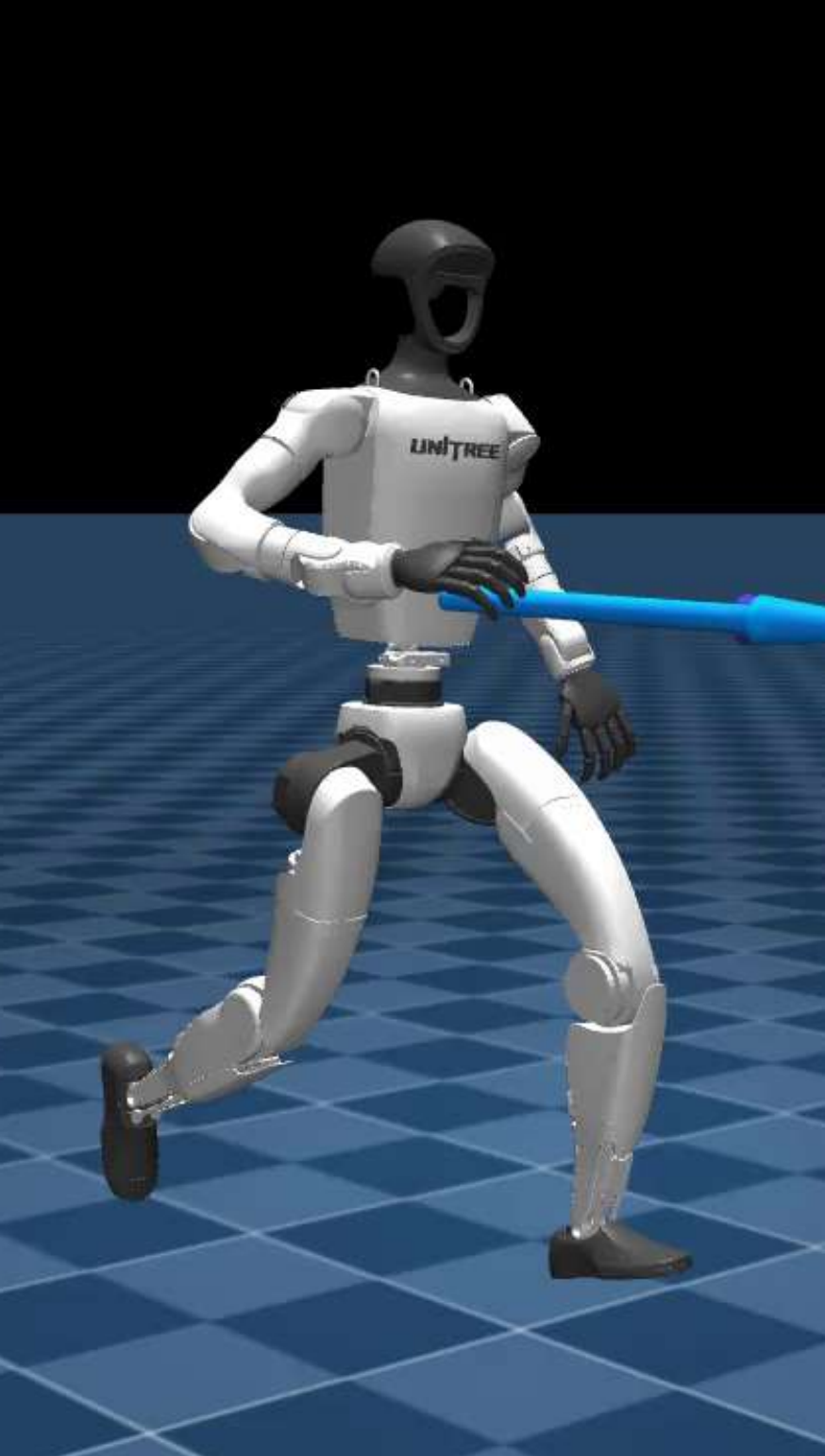}
        \caption{$5$}
    \end{subfigure}\hfill
    \begin{subfigure}[t]{0.19\columnwidth}
        \centering
        \includegraphics[width=\linewidth,trim={26.46cm 6.00cm 26.46cm 6.00cm},clip]{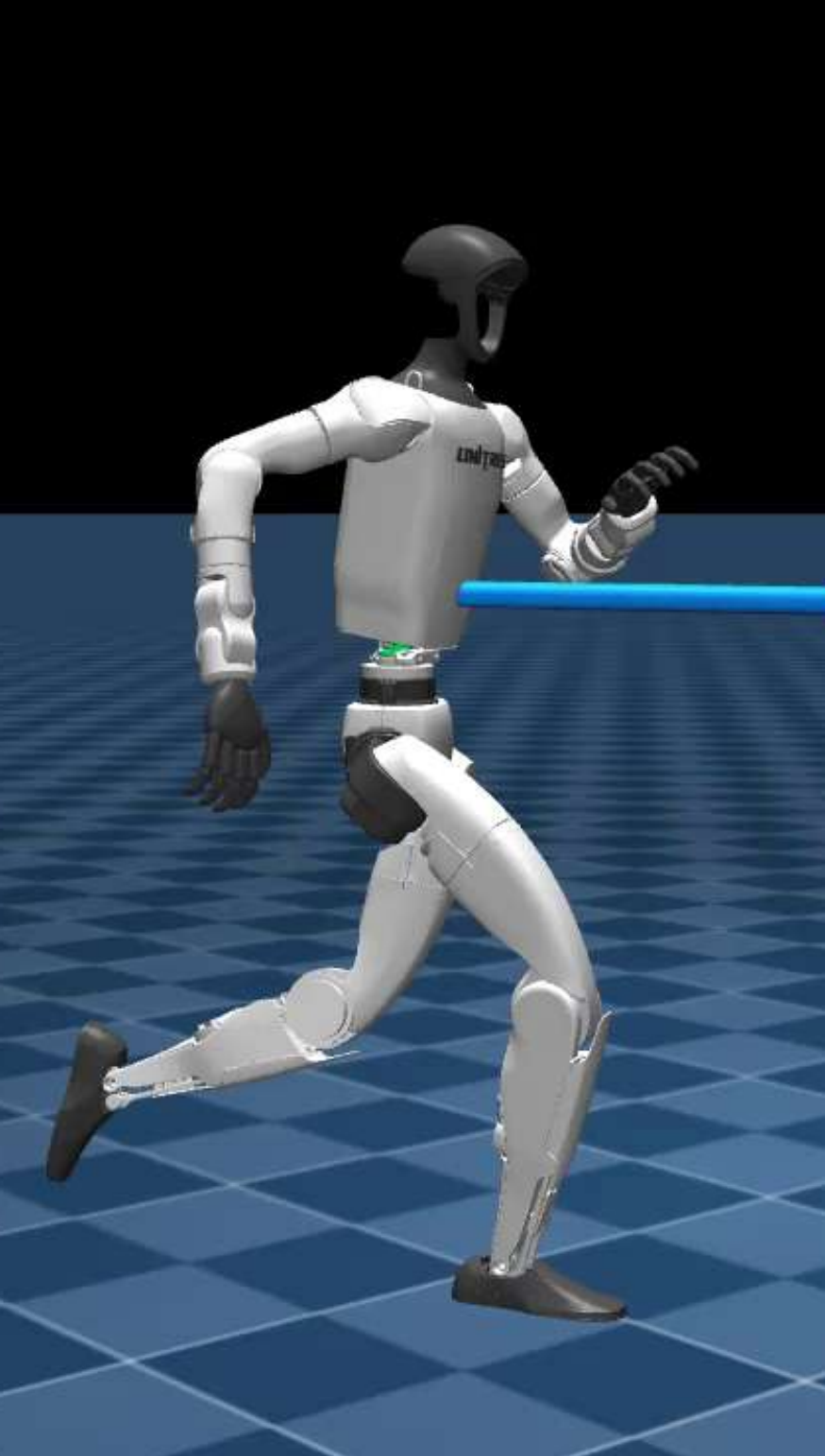}
        \caption{$8$}
    \end{subfigure}
    \caption{Representative gaits generated by task policies with different
    active RVQ depths $M_{\mathrm{act}}$ (shown below each frame). One
    active stage mainly produces standing behavior; two stages produce a slow
    walking gait; three stages enable fast walking; five or more stages produce
    faster and more natural running.}
    \label{fig:rvq_depth_gaits}
\end{figure}

\begin{table}[!ht]
\centering
\caption{Forward-speed progression at different active RVQ depths.
Columns $\bar{v}_{0.5}$, $\bar{v}_{1.5}$, and $\bar{v}_{3.0}$ report achieved
forward speed along the target heading for the corresponding commanded speed.
RMSE, fall rate, and action rate are reported at
$v^\star=3.0\,\mathrm{m\,s^{-1}}$.}
\label{tab:rvq_depth_locomotion}
\small
\setlength{\tabcolsep}{1.5pt}
\begin{tabular}{@{}ccccccc@{}}
\toprule
$M_{\mathrm{act}}$ &
$\bar{v}_{0.5}$ &
$\bar{v}_{1.5}$ &
$\bar{v}_{3.0}$ &
RMSE$_{3.0}$ &
Falls$_{3.0}$ &
Act. rate$_{3.0}$ \\
\midrule
1 & $0.02$ & $0.17$ & $0.00$ & $3.00$ & $0.2\%$ & $0.039$ \\
2 & $0.22$ & $0.40$ & $0.07$ & $2.94$ & $0.1\%$ & $0.031$ \\
3 & $0.37$ & $1.43$ & $2.88$ & $0.21$ & $0.0\%$ & $0.102$ \\
5 & $0.49$ & $1.52$ & $2.95$ & $0.18$ & $0.0\%$ & $0.092$ \\
8 & $0.53$ & $1.47$ & $2.87$ & $0.20$ & $0.0\%$ & $0.130$ \\
\bottomrule
\end{tabular}
\end{table}

The first two depths are stable but do not provide enough discrete modulation
to generate fast locomotion. With one active stage, the policy remains
essentially stationary even under high forward-speed commands. With two stages,
a slow walking pattern appears, but the achieved speed saturates far below the
commanded values. The main transition occurs at three active stages: the policy
reaches fast-walking, almost-running behavior and follows the progression up to
$3.0\,\mathrm{m\,s^{-1}}$ without falls.

With additional stages, the policy develops faster and more natural running, as
shown qualitatively in Fig.~\ref{fig:rvq_depth_gaits}. At full depth, the
policy achieves slightly worse tracking and a higher action rate, suggesting
that $M_{\mathrm{act}}=5$ provides the best trade-off for this task. Overall,
the progression suggests an ordered structure in the
RVQ hierarchy: early stages are sufficient for balance and slow walking,
while deeper stages enable faster gaits and more dynamic motions.

\subsection{Effect of the Rotation Trick on Codebook Structure}
\label{sec:rotation_trick}

To isolate the contribution of the rotation trick, we train a second distilled
HMP using the standard straight-through estimator (STE), while keeping the
architecture, expert data, and number of training iterations unchanged.
We refer to the two variants as \emph{RT} (ours) and \emph{STE}. For the latent-space analysis,
both variants are evaluated with their corresponding reference encoders on the
same $30$\,s expert rollouts generated in $1{,}024$ parallel environments.
Converged training losses are averaged over the last $500$ distillation
iterations.

\subsubsection{Imitation Fidelity}

Both HMPs reach comparable action-level MSE against the expert
($1.3\!\times\!10^{-3}$ for RT and $1.2\!\times\!10^{-3}$ for STE); we
therefore focus on differences in latent organization.

\subsubsection{Latent-Space Decomposition}
Table~\ref{tab:rt_vs_ste_latent} summarizes the relevant quantities. Without
the rotation trick, the codebook contribution, residual magnitude, and
quantization error are all more than twice their RT counterparts. In the STE variant, the reference encoder
expresses a larger share of the action-generating latent through the quantized
residual, instead of relying on the continuous proprioceptive latent. Converged training losses
show the same pattern, remaining
$3.8$--$5.9\times$ higher for STE.

\subsubsection{Code Utilization}
The rotation trick is known to improve code utilization~\cite{fifty2025rotation},
and Table~\ref{tab:rt_vs_ste_latent} confirms the same trend here: the STE
variant exhibits lower global perplexity across stages, indicating a less
uniform use of the available codes.

\subsubsection{Stage-Wise Residuals}
Following the hierarchy observed in Sec.~\ref{sec:depth}, we group the first
three quantization stages as coarse and the remaining five as fine.
Table~\ref{tab:rt_vs_ste_latent} shows that the difference between STE and RT
persists across the codebook hierarchy. The ratio between STE and RT remains
essentially unchanged for the residual after the coarse stages and the
subsequent fine-stage refinement, showing that the effect of the rotation trick persists across the full RVQ depth.

\begin{table}[!ht]
\centering
\caption{Rotation trick (ours) vs.\ standard straight-through estimator (STE).
Upper block: latent-space quantities on expert rollouts in $1{,}024$ parallel
environments.
Middle block: magnitude of the residual after the coarse stages (1--3) and of
the subsequent refinement introduced by the fine stages (4--8).
Lower block: converged training losses, averaged over the last $500$
distillation iterations.
}
\label{tab:rt_vs_ste_latent}
\small
\setlength{\tabcolsep}{3pt}
\begin{tabular}{@{}lccc@{}}
\toprule
Quantity & RT (ours) & STE & Ratio \\
\midrule
\multicolumn{4}{@{}l}{\emph{Latent-Space Quantities}} \\
Codebook magnitude $\|\hat{y}_t\|$        & $\mathbf{0.44}$ & $0.97$ & $2.2\times$ \\
Residual magnitude $\|y_t\|$              & $\mathbf{0.46}$ & $1.01$ & $2.2\times$ \\
Quant.\ error $\|y_t-\hat{y}_t\|$         & $\mathbf{0.14}$ & $0.34$ & $2.4\times$ \\
Global perplexity (avg.\ stages)          & $\mathbf{728}$  & $684$  & --- \\
\midrule
\multicolumn{4}{@{}l}{\emph{Stage-Wise Residuals}} \\
After coarse stages (1--3) & $\mathbf{0.31}$ & $0.71$ & $2.3\times$ \\
Fine-stage refinement (4--8) & $\mathbf{0.18}$ & $0.43$ & $2.4\times$ \\
\midrule
\multicolumn{4}{@{}l}{\emph{Training Losses}} \\
$\mathcal{L}_m$ (magnitude)      & $\mathbf{1.3\!\times\!10^{-3}}$ & $6.9\!\times\!10^{-3}$ & $5.2\times$ \\
$\mathcal{L}_r$ (temporal reg.)  & $\mathbf{5.1\!\times\!10^{-3}}$ & $1.9\!\times\!10^{-2}$ & $3.8\times$ \\
$\mathcal{L}_q$ (commitment)     & $\mathbf{7.9\!\times\!10^{-4}}$ & $4.7\!\times\!10^{-3}$ & $5.9\times$ \\
\bottomrule
\end{tabular}
\end{table}

\subsection{Rotation Trick and Downstream Robustness}
\label{sec:velocity_tracking}

Finally, we train task-level velocity-tracking policies on the RT and STE HMPs
(identical command distribution, reward, PPO hyperparameters,
$M_{\mathrm{act}}=8$) and evaluate each over $1{,}024$ simulation rollouts of
$30$\,s under the final command range. Results are reported in Table~\ref{tab:velocity_tracking}.

\begin{table}[!ht]
\centering
\caption{Velocity tracking at $M_{\mathrm{act}}=8$ over $1{,}024$ rollouts of
$30$\,s under the final command range.}
\label{tab:velocity_tracking}
\small
\setlength{\tabcolsep}{4pt}
\begin{tabular}{@{}lccc@{}}
\toprule
Model & Lin.\ RMSE $[\mathrm{m\,s^{-1}}]$ & Ang.\ RMSE $[\mathrm{rad\,s^{-1}}]$ & Fall rate \\
\midrule
RT (ours) & $\mathbf{0.46\pm0.15}$ & $1.03\pm0.46$          & $\mathbf{2.1\%}$ \\
STE       & $0.49\pm0.18$          & $\mathbf{0.95\pm0.42}$ & $10.2\%$ \\
\bottomrule
\end{tabular}
\end{table}

The two variants achieve similar tracking errors, but differ strongly in
robustness. STE falls about five times more often than RT ($10.2\%$ vs.\
$2.1\%$). This suggests that the representation
learned with the rotation trick is less prone to unstable downstream behavior
under challenging velocity commands.

\section{Conclusion}

We presented a three-stage pipeline for reusing motion-imitation skills in
humanoid locomotion. A reference-conditioned expert is first trained from
retargeted human motions, then distilled into a frozen hybrid motion prior (HMP)
composed of a proprioceptive encoder, RVQ codebook, and action decoder.
Downstream task-level policies are finally trained to solve locomotion tasks by
selecting discrete codebook entries, without modifying the previously acquired low-level motor skills.

The distillation evaluation shows that the reference encoder and HMP together
preserve the tracking behavior of the expert. At deployment, the reference
encoder is removed and task-level policies directly select codebook entries
through the frozen HMP.
On the Unitree G1, task-level policies trained with the same frozen HMP solve
velocity tracking, point-goal navigation, and fall-recovery velocity tracking
in simulation, and the velocity-tracking policy transfers to the real robot
without real-world fine-tuning. The analysis further shows that the
proprioceptive encoder and the codebook play complementary roles: the continuous
latent carries balance and basic locomotion competence, while the codebook
provides discrete task-conditioned modulations. Varying the active RVQ depth reveals an
ordered set of standing, walking, fast walking, and running behaviors, and
training the codebook with the rotation trick improves latent
organization and downstream robustness compared with a standard
straight-through estimator.

Some limitations remain. In hardware, the policy is balanced and responsive to
commands, but the resulting motions still exhibit visible oscillations during
walking and turning. Future work will extend the evaluation to a broader set of
downstream tasks, including loco-manipulation, and study how to improve the
smoothness and robustness of the frozen HMP under more dynamic real-world
motions.

\IEEEtriggeratref{5}
\bibliographystyle{IEEEtran}
\bibliography{references}

\begin{thebibliography}{10}
\providecommand{\url}[1]{#1}
\csname url@samestyle\endcsname
\providecommand{\newblock}{\relax}
\providecommand{\bibinfo}[2]{#2}
\providecommand{\BIBentrySTDinterwordspacing}{\spaceskip=0pt\relax}
\providecommand{\BIBentryALTinterwordstretchfactor}{4}
\providecommand{\BIBentryALTinterwordspacing}{\spaceskip=\fontdimen2\font plus
\BIBentryALTinterwordstretchfactor\fontdimen3\font minus \fontdimen4\font\relax}
\providecommand{\BIBforeignlanguage}[2]{{%
\expandafter\ifx\csname l@#1\endcsname\relax
\typeout{** WARNING: IEEEtran.bst: No hyphenation pattern has been}%
\typeout{** loaded for the language `#1'. Using the pattern for}%
\typeout{** the default language instead.}%
\else
\language=\csname l@#1\endcsname
\fi
#2}}
\providecommand{\BIBdecl}{\relax}
\BIBdecl

\bibitem{rudin2022walk}
N.~Rudin, D.~Hoeller, P.~Reist, and M.~Hutter, ``Learning to walk in minutes using massively parallel deep reinforcement learning,'' in \emph{Conference on Robot Learning (CoRL)}.\hskip 1em plus 0.5em minus 0.4em\relax PMLR, 2021, pp. 91--100.

\bibitem{radosavovic2024real}
I.~Radosavovic, T.~Xiao, B.~Zhang, T.~Darrell, J.~Malik, and K.~Sreenath, ``Real-world humanoid locomotion with reinforcement learning,'' \emph{Science Robotics}, vol.~9, no.~89, p. eadi9579, 2024.

\bibitem{cheng2024expressive}
X.~Cheng, Y.~Ji, J.~Chen, R.~Yang, G.~Yang, and X.~Wang, ``Expressive whole-body control for humanoid robots,'' in \emph{Robotics: Science and Systems (RSS)}, 2024.

\bibitem{fu2024humanplus}
Z.~Fu, Q.~Zhao, Q.~Wu, G.~Wetzstein, and C.~Finn, ``Humanplus: Humanoid shadowing and imitation from humans,'' \emph{arXiv preprint arXiv:2406.10454}, 2024.

\bibitem{peng2018deepmimic}
X.~B. Peng, P.~Abbeel, S.~Levine, and M.~Van~de Panne, ``Deepmimic: Example-guided deep reinforcement learning of physics-based character skills,'' \emph{ACM Transactions on Graphics (TOG)}, vol.~37, no.~4, pp. 1--14, 2018.

\bibitem{liao2025beyondmimic}
Q.~Liao, T.~E. Truong, X.~Huang, G.~Tevet, K.~Sreenath, and C.~K. Liu, ``Beyondmimic: From motion tracking to versatile humanoid control via guided diffusion,'' \emph{arXiv preprint arXiv:2508.08241}, 2025.

\bibitem{fifty2025rotation}
C.~Fifty, R.~G. Junkins, D.~Duan, A.~Iyengar, J.~W. Liu, E.~Amid, S.~Thrun, and C.~R{\'e}, ``Restructuring vector quantization with the rotation trick,'' in \emph{International Conference on Learning Representations (ICLR)}, 2025, oral presentation. arXiv:2410.06424.

\bibitem{luo2023phc}
Z.~Luo, J.~Cao, K.~Kitani, and W.~Xu, ``Perpetual humanoid control for real-time simulated avatars,'' in \emph{Proceedings of the IEEE/CVF International Conference on Computer Vision (ICCV)}, 2023, pp. 10\,895--10\,904.

\bibitem{merel2019npmp}
J.~Merel, L.~Hasenclever, A.~Galashov, A.~Ahuja, V.~Pham, G.~Wayne, Y.~W. Teh, and N.~Heess, ``Neural probabilistic motor primitives for humanoid control,'' in \emph{International Conference on Learning Representations (ICLR)}, 2019.

\bibitem{peng2022ase}
X.~B. Peng, Y.~Guo, L.~Halper, S.~Levine, and S.~Fidler, ``Ase: Large-scale reusable adversarial skill embeddings for physically simulated characters,'' \emph{ACM Transactions on Graphics (TOG)}, vol.~41, no.~4, 2022.

\bibitem{luo2024pulse}
Z.~Luo, J.~Cao, J.~Merel, A.~Winkler, J.~Huang, K.~M. Kitani, and W.~Xu, ``Universal humanoid motion representations for physics-based control,'' in \emph{International Conference on Learning Representations (ICLR)}, 2024.

\bibitem{11203023}
M.~Stępień, R.~Kourdis, C.~Roux, and O.~Stasse, ``Latent conditioned loco-manipulation using motion priors,'' in \emph{2025 IEEE-RAS 24th International Conference on Humanoid Robots (Humanoids)}, 2025, pp. 365--372.

\bibitem{11244136}
C.~Tsakonas and K.~Chatzilygeroudis, ``Vector quantized-elites: Unsupervised and problem-agnostic quality-diversity optimization,'' \emph{IEEE Transactions on Evolutionary Computation}, pp. 1--1, 2025.

\bibitem{vandenoord2017vqvae}
A.~van~den Oord, O.~Vinyals, and K.~Kavukcuoglu, ``Neural discrete representation learning,'' in \emph{Advances in Neural Information Processing Systems (NeurIPS)}, 2017.

\bibitem{lee2022rqvae}
D.~Lee, C.~Kim, S.~Kim, M.~Cho, and W.-S. Han, ``Autoregressive image generation using residual quantization,'' in \emph{Proceedings of the IEEE/CVF Conference on Computer Vision and Pattern Recognition (CVPR)}, 2022, pp. 11\,523--11\,532.

\bibitem{zhu2023ncp}
Q.~Zhu, H.~Zhang, M.~Lan, and L.~Han, ``Neural categorical priors for physics-based character control,'' \emph{ACM Transactions on Graphics (TOG)}, vol.~42, no.~6, pp. 1--16, 2023.

\bibitem{bae2025hybrid}
J.~Bae, J.~Won, D.~Lim, I.~Hwang, and Y.~M. Kim, ``Versatile physics-based character control with hybrid latent representation,'' \emph{Computer Graphics Forum}, vol.~44, no.~2, p. e70018, 2025, eurographics 2025.

\bibitem{wu2025discretepolicy}
K.~Wu, Y.~Zhu, J.~Li, J.~Wen, N.~Liu, Z.~Xu, Q.~Qiu, and J.~Tang, ``Discrete policy: Learning disentangled action space for multi-task robotic manipulation,'' in \emph{IEEE International Conference on Robotics and Automation (ICRA)}, 2025.

\bibitem{amadio2026yahmp}
F.~Amadio and E.~Mingo~Hoffman, ``What matters in humanoid general motion tracking? an empirical study,'' \emph{arXiv preprint arXiv:2607.19903}, 2026.

\bibitem{schulman2017proximal}
\BIBentryALTinterwordspacing
J.~Schulman, F.~Wolski, P.~Dhariwal, A.~Radford, and O.~Klimov, ``Proximal policy optimization algorithms,'' \emph{CoRR}, vol. abs/1707.06347, 2017. [Online]. Available: \url{http://arxiv.org/abs/1707.06347}
\BIBentrySTDinterwordspacing

\bibitem{li2025reinforcement}
Z.~Li, X.~B. Peng, P.~Abbeel, S.~Levine, G.~Berseth, and K.~Sreenath, ``Reinforcement learning for versatile, dynamic, and robust bipedal locomotion control,'' \emph{The International Journal of Robotics Research}, vol.~44, no.~5, pp. 840--888, 2025.

\bibitem{todorov2012mujoco}
E.~Todorov, T.~Erez, and Y.~Tassa, ``Mujoco: A physics engine for model-based control,'' in \emph{2012 IEEE/RSJ International Conference on Intelligent Robots and Systems (IROS)}.\hskip 1em plus 0.5em minus 0.4em\relax IEEE, 2012, pp. 5026--5033.

\bibitem{zakka2026mjlab}
K.~Zakka, Q.~Liao, B.~Yi, L.~L. Lay, K.~Sreenath, and P.~Abbeel, ``mjlab: A lightweight framework for gpu-accelerated robot learning,'' 2026.

\bibitem{harvey2020robust}
F.~G. Harvey, M.~Yurick, D.~Nowrouzezahrai, and C.~Pal, ``Robust motion in-betweening,'' \emph{ACM Transactions on Graphics (TOG)}, vol.~39, no.~4, pp. 60--1, 2020.

\end{thebibliography}

\end{document}